\documentclass[]{template}

\usepackage[utf8]{inputenc}             %
\usepackage[T1]{fontenc}                %
\usepackage{url}                        %
\usepackage{booktabs}                   %
\usepackage{multirow}
\usepackage{colortbl}
\usepackage{multicol}
\usepackage{amsfonts}                   %
\usepackage{nicefrac}                   %
\usepackage{microtype}                  %
\usepackage{times}
\usepackage{inconsolata}
\usepackage[dvipsnames]{xcolor}         %

\usepackage{latexsym}

\usepackage{graphicx}
\usepackage{float}
\usepackage{subcaption}
\usepackage{wrapfig}
\usepackage{lipsum}

\usepackage{siunitx}
\usepackage{tablefootnote}
\usepackage{makecell}

\usepackage{bm}

\usepackage{tabularx} 
\usepackage{ragged2e} 
\newcolumntype{L}{>{\RaggedRight\hangafter=1\hangindent=0em}X}

\usepackage{enumitem}

\usepackage{amsmath}
\usepackage{amssymb}
\usepackage{mathtools}
\usepackage{amsthm}

\setboolean{logo}{true}    %

\usepackage[linesnumbered,ruled,vlined]{algorithm2e}

\hypersetup{
    colorlinks=true,
    linkcolor=red,
    citecolor=Cerulean,
    filecolor=magenta,      
    urlcolor=magenta,
}

\usepackage[capitalize,noabbrev]{cleveref}
\crefname{section}{§}{§§}
\Crefname{section}{§}{§§}

\usepackage{calligra}
\DeclareMathAlphabet{\mathcalligra}{T1}{calligra}{m}{n}

\usepackage{pifont}

\theoremstyle{plain}

\theoremstyle{definition}

\theoremstyle{remark}

\renewcommand{\paragraph}[1]{\vspace{1mm}\noindent\textbf{#1}\hspace{1em}}

\DeclareCaptionLabelFormat{cont}{#1~#2\alph{ContinuedFloat}}
\usepackage[most]{tcolorbox}

\tcbset{
  promptbox/.style={
    enhanced,
    breakable,
    colback=white,
    colframe=black!50,
    boxrule=0.6pt,
    left=10pt,right=10pt,top=1.0mm,bottom=1.0mm,
    fonttitle=\bfseries,
    title=#1,
  }
}
\newtcolorbox{promptbox}[1]{promptbox={#1}}

\tcbset{
  takeawaybox/.style={
    top=10pt,
    colback=lightgray!20,
    colframe=Black,
    colbacktitle=BurntOrange,
    enhanced,
    center,
    attach boxed title to top center={yshift=-0.1in,xshift=0.0in},
    boxed title style={boxrule=0pt,colframe=white,},
  }
}
\newtcolorbox{takeawaybox}[2][]{takeawaybox, title=#2,#1}
\tcbset{
  observationbox/.style={
    top=10pt,
    colback=lightgray!20,
    colframe=Black,
    colbacktitle=YellowGreen,
    enhanced,
    center,
    attach boxed title to top center={yshift=-0.1in,xshift=0.0in},
    boxed title style={boxrule=0pt,colframe=white,},
  }
}
\newtcolorbox{observationbox}[2][]{observationbox, title=#2,#1}

\usepackage{xfp}
\newcommand{\TokPercent}[1]{
  \fpeval{min(55, max(0, (#1)/1.8 * 55))}
}

\newcommand{\Tok}[5]{%
  \begingroup
  \setlength{\fboxsep}{1.3pt}%
  \renewcommand{\arraystretch}{0.82}%
  \setlength{\tabcolsep}{0pt}%
  \begin{tabular}[t]{@{}c@{}}%
    \colorbox{red!\TokPercent{#4}}{\strut\kern1.2pt#1\kern1.2pt}\\[-0.25ex]
    {\tiny\color{black!70}Rank:~#2}\\[-0.25ex]
    {\tiny\color{black!70}Surprisal:~#3}\\[-0.25ex]
    {\tiny\color{black!70}\RatioShort$^*_{\text{token}}$:~#5}%
  \end{tabular}%
  \hspace{0.2em}
  \endgroup
}

\newcommand{\TokEllipsis}{%
  \begingroup
  \setlength{\fboxsep}{1.3pt}
  \renewcommand{\arraystretch}{0.82}
  \setlength{\tabcolsep}{0pt}
  \begin{tabular}[t]{@{}c@{}}
    \strut\kern1.2pt...\kern1.2pt\\[-0.25ex]
    {\tiny\phantom{rank: 000}}\\[-0.25ex]
    {\tiny\phantom{nll: 00.00}}\\[-0.25ex]
    {\tiny\phantom{\RatioShort$^*_{\text{token}}$: 00.00}}
  \end{tabular}
  \endgroup
}

\newcommand\blfootnote[1]{%
  \begingroup
  \renewcommand\thefootnote{}\footnote{#1}%
  \addtocounter{footnote}{-1}%
  \endgroup
}

\usepackage{CJK}
\usepackage{xspace}
\newcommand{\RatioShort}{RSR}
\newcommand{\RatioShortNew}{RSR\xspace}
\newcommand{\Ratio}{\emph{Rank-Surprisal Ratio}\xspace}

\newcommand{\sci}[2]{#1$_{\times 10^{#2}}$}

\colorlet{RankColorBase}{blue!55!gray}
\newcommand{\RankShade}[1]{%
  \ifcase#1\relax
  \or\cellcolor{RankColorBase!20}%
  \or\cellcolor{RankColorBase!16}%
  \or\cellcolor{RankColorBase!12}%
  \or\cellcolor{RankColorBase!8}%
  \or\cellcolor{RankColorBase!4}%
  \or\cellcolor{RankColorBase!0}%
  \fi
}
\newcommand{\best}[1]{\cellcolor{RankColorBase!20}{#1}}
\newcommand{\second}[1]{\cellcolor{RankColorBase!8}{#1}}

\title{Which Reasoning Trajectories Teach Students to Reason Better? A Simple Metric of Informative Alignment}

\author[1,2]{Yuming Yang}
\author[3]{Mingyoung Lai}
\author[1]{Wanxu Zhao}
\author[1]{Xiaoran Fan}
\author[1]{Zhiheng Xi}
\author[1]{Mingqi Wu}
\author[4]{Chiyue Huang}
\author[1]{Jun Zhao}
\author[2]{Haijun Lv}
\author[2]{Jian Tong}
\author[2]{Yunhua Zhou}
\author[2,$\dagger$]{Yicheng Zou}
\author[2]{Qipeng Guo}
\author[1]{Tao Gui}
\author[1,$\dagger$]{Qi Zhang}
\author[1]{Xuanjing Huang}

\affil[1]{Fudan University}
\affil[2]{Shanghai AI Laboratory}
\affil[3]{University of Toronto}
\affil[4]{University of Sydney}

\begin{abstract}
Long chain-of-thought (CoT) trajectories provide rich supervision signals for distilling reasoning from teacher to student LLMs.
However, both prior work and our experiments show that trajectories from stronger teachers do not necessarily yield better students, highlighting the importance of data-student suitability in distillation.
Existing methods assess suitability primarily through student likelihood, favoring trajectories that align closely with the student model's current behavior but overlooking more informative ones.
Addressing this, we propose \Ratio\ (\RatioShort), a simple metric that captures both alignment and informativeness to assess the suitability of a reasoning trajectory.
\RatioShort\ is motivated by the observation that effective trajectories typically balance learning signal strength and behavioral alignment by combining low absolute probability with relatively high-ranked tokens under the student model.
Concretely, \RatioShort\ is defined as the ratio of a trajectory's average token-wise rank to its average negative log-likelihood, and is straightforward to compute and interpret.
Across five student models and reasoning trajectories from 11 diverse teachers, \RatioShort\ strongly correlates with post-training reasoning performance (average Spearman 0.86), consistently outperforming existing metrics. 
We further demonstrate its practical utility in both trajectory selection and teacher selection.

\end{abstract}

\begin{document}

\blfootnote{$\dagger$ Corresponding authors. Inquiries may be sent to: yumingyang23@m.fudan.edu.cn, zouyicheng@pjlab.org.cn, qz@fudan.edu.cn}
\blfootnote{$*$ Code and data are available at \url{https://github.com/UmeanNever/RankSurprisalRatio}.}

\maketitle

\section{Introduction}

Recent advances in reasoning-oriented large language models (LLMs) are largely driven by their ability to generate long chain-of-thought (CoT) trajectories \cite{wei2022chain, DBLP:journals/corr/abs-2503-24235}.
Beyond enabling complex inference at test time, such trajectories also provide powerful supervision signals for training student models \cite{DeepseekR1, DBLP:journals/corr/abs-2501-19393} or cold-starting reinforcement learning \cite{Qwen3} through supervised fine-tuning (SFT).

Yet, stronger reasoning teachers do not necessarily yield better students \cite{SmallModels,OpenThoughts}. 
Our extensive experiments show that the post-training effectiveness of reasoning trajectories varies substantially across student models, indicating that the suitability between data and student is critical for effective learning.
Existing data engineering methods assess data suitability primarily through the student’s probability assignments \cite{DBLP:journals/corr/abs-2502-04194, DBLP:journals/corr/abs-2510-03988}, favoring high-likelihood trajectories that align closely with the model's current behavior.
Such trajectories, however, often provide limited new learning signals.
In contrast, more informative trajectories are typically less familiar to the student and thus overlooked by these methods.
To facilitate more effective learning, it is crucial to strike a balance between familiarity and informativeness, echoing the psychological concept of the zone of proximal development \cite{vygotsky1978mind}.
This leads to a fundamental \emph{Informative Alignment} challenge: \textbf{how to identify reasoning data that are both well aligned with the student and sufficiently informative?}

\begin{wrapfigure}{r}{0.48\linewidth}
    \centering
    \vspace{-1em}
        \includegraphics[width=\linewidth]{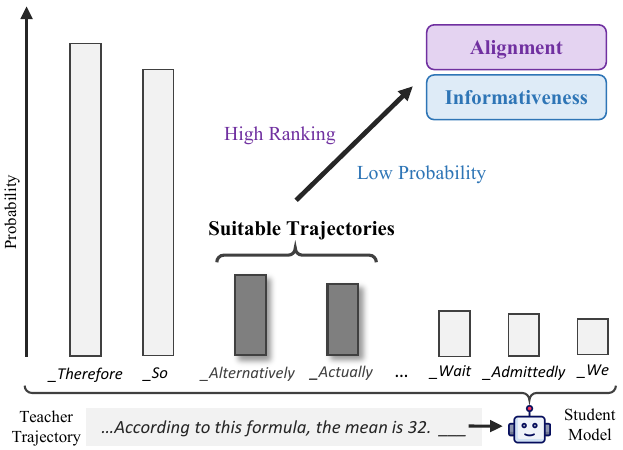}
    \caption{Illustration of the intuition behind \Ratio. Suitable reasoning trajectories should balance informativeness and alignment, having low absolute probability while their tokens remain relatively high-ranked under the student model.}
    \label{fig:head}
    \vspace{-1em}
\end{wrapfigure}

To address this challenge, we propose a simple yet effective metric, \Ratio\ (\RatioShort), which quantifies the suitability of a reasoning trajectory for a given student by jointly capturing alignment and informativeness.
Motivated by our preliminary analysis, we argue that the dilemma between providing new signals and aligning with student's existing behavior can be resolved by trajectories exhibiting both absolute unfamiliarity and relative familiarity.
Concretely, effective trajectories should deviate from the student's own generations, receiving \textbf{low absolute probability} under the student model, while remaining compatible with its overall generation patterns, such that their tokens still \textbf{rank relatively high} in the model’s prediction distribution over the vocabulary (Figure~\ref{fig:head}).

Based on this insight and consistent numerical patterns observed in simulation studies, we define our suitability metric, \Ratio, as the ratio between a trajectory’s average token-wise rank\footnote{Higher-ranked tokens have lower rank values.} and its average negative log-likelihood (surprisal). 
\RatioShort\ can be computed with a single forward pass, requires no additional verifier or test data, and is straightforward to interpret.
Lower \RatioShort\ indicates better informative alignment, identifying trajectories that are both informative and well aligned with the student.

We validate the effectiveness of \Ratio\ through correlation analyses on 5 student LLMs using math reasoning trajectories generated by 11 representative teacher models.
Across all students, the \RatioShort\ of trajectories exhibits a strong correlation with post-training performance, achieving an average Spearman correlation of 0.86 and consistently outperforming alternative metrics.
Furthermore, to explore its practical value in data engineering, we apply \RatioShort\ to trajectory selection and teacher selection. Our experiments show that \RatioShort\ not only selects more effective training trajectories for each problem from candidates generated by diverse teachers, but also identifies more suitable teacher models using only a small amount of data, consistently outperforming existing selection methods across all five students in both settings.

Our main contributions are three-fold:
\begin{itemize}[itemsep=3pt, parsep=0pt, topsep=3pt]
\item We present a systematic distillation study across a wide range of teacher and student models, showing that the effectiveness of reasoning trajectories differs across students and highlighting the importance of data-student suitability (\S~\ref{sec:pre}).
\item We propose \Ratio, a simple metric that quantifies the suitability of a reasoning trajectory for a given student model by jointly capturing alignment and informativeness (\S~\ref{sec:metric}), achieving a strong correlation with post-training performance (\S~\ref{sec:corr}).
\item We demonstrate the practical utility of \RatioShortNew in two data engineering scenarios, trajectory selection and teacher selection, where it serves as an effective criterion and outperforms existing methods (\S~\ref{sec:select}).
We further discuss the broader applicability of \RatioShortNew to non-CoT data and subset selection (\S~\ref{sec:broad}).
\end{itemize}

\section{The Need for Student-Specific Data}
\label{sec:pre}

To understand which types of reasoning trajectories most effectively improve student models after SFT, we conduct a comprehensive large-scale study involving five widely adopted student models and eleven diverse reasoning-oriented teacher models, yielding 55 teacher-student pairings.
We perform SFT experiments for each of these pairs.

\begin{table*}[t!]
\centering
\resizebox{\linewidth}{!}{
\begin{tabular}{lccccccc}
\toprule
\multirow{2}{*}{\textbf{Teacher Models}} & \multirow{2}{*}{\textbf{Params}} &
\multicolumn{5}{c}{\textbf{Student Models (Base)}} & \multirow{2}{*}{\textbf{\makecell{Teacher\\Performance}}}\\
\cmidrule(lr){3-7}
& & \textbf{Qwen-3-14B} & \textbf{LLaMA-3.1-8B} & \textbf{Qwen-2.5-7B} & \textbf{Qwen-3-4B} & \textbf{Qwen-2.5-3B} & \\
\midrule
Deepseek-R1              & 671B & 77.1 & \second{28.1} & 47.3 & 55.8 & 29.6 & 91.1 \\
Qwen-3-235B-Thinking      & 235B & 71.8 & 22.0 & 45.0 & 53.4 & 26.4 & \second{91.2} \\
GPT-OSS-120B              & 120B & 66.7 & 15.2 & 40.7 & 47.9 & 22.9 & 88.3 \\
Nemotron-Super            & 49B  & 72.2 & 23.7 & 48.3 & 56.4 & 33.0 & 82.3 \\
QwQ-32B                   & 32B  & \best{77.4} & 27.1 & \best{52.0} & 61.2 & 33.0 & 85.2 \\
Qwen-3-30B-Thinking       & 30B  & \second{77.2} & 26.7 & 50.0 & 58.8 & 31.2 & \best{92.3} \\
Magistral-Small           & 24B  & 68.8 & 22.8 & 47.6 & 52.2 & 30.6 & 71.0 \\
GPT-OSS-20B               & 20B  & 69.5 & 17.9 & 42.7 & 48.4 & 24.4 & 83.4 \\
Phi-4-Reasoning-Plus      & 14B  & 54.1 & 14.5 & 35.2 & 40.2 & 18.2 & 72.7 \\
Qwen-3-8B                 & 8B   & 74.6 & 26.5 & \second{52.0} & \second{61.2} & \best{34.2} & 82.5 \\
Qwen-3-4B-Thinking        & 4B   & 76.8 & \best{28.2} & 51.8 & \best{61.9} & \second{33.3} & 87.3 \\
\bottomrule
\end{tabular}
}
\caption{Distillation results showing post-training reasoning performance of student models trained on trajectories from different teacher models, evaluated by average Acc@4 on AIME’25, AIME’24, AMC’23, and MATH500. Darker and lighter shading indicate the best and second-best results, respectively. Student performance varies significantly across teacher-student pairs, highlighting the importance of data-student suitability.}
\label{tab:teacher_student_scores}
\vspace{-1.0em}
\end{table*}

\subsection{Experimental Settings}
\label{subsec:exp_setting}

Our teacher-student pairing study involves two major steps:
(1) For each teacher model, we prompt it to generate a long CoT response for each math problem in our 5000-problem set (see \S~\ref{app:problemset}), forming a trajectory dataset specific to that teacher.
(2) For each teacher-student pair, we fine-tune the student model on the corresponding teacher dataset and evaluate its reasoning performance. 
All students are pre-trained base models.
To reduce variance induced by stochastic trajectory sampling, we perform three independent generation runs for each teacher and conduct SFT separately on each resulting dataset for every teacher-student pair. Reported results are averaged over these three runs.
More implementation details are provided in Appendix~\ref{app:exp}.

\paragraph{Teachers}
We use 11 reasoning LLMs (\S~\ref{app:teachers}) spanning 4B to 671B parameters across multiple model families, including DeepSeek \cite{DeepseekR1}, GPT-OSS \cite{agarwal2025gpt}, Qwen \cite{Qwen3}, LLaMA-Nemotron \cite{bercovich2025llama}, and Phi \cite{abdin2025phi}.

\paragraph{Benchmarks}
We evaluate the reasoning performance of fine-tuned student models on four popular math benchmarks (AIME'25, AIME'24, AMC'23, and MATH500 \cite{hendrycks2021measuring}) using the Acc@4 metric (\S~\ref{app:bench}), and report results averaged across all benchmarks.

\subsection{Results}
\label{subsec:teacher_student_res}

Table~\ref{tab:teacher_student_scores} presents the results of our teacher-student pairing distillation study, revealing that:

\paragraph{Stronger teachers do not necessarily produce better students.} 
Teacher capability, whether measured by parameter scale or reasoning performance, does not reliably predict student improvement. For example, the 671B and 235B models often underperform smaller teachers such as QwQ-32B on multiple students. Similarly, teachers with strong reasoning performance do not consistently yield the best outcomes for all student models.

\paragraph{Data-student suitability is critical for eliciting reasoning improvements.}
The effectiveness of teacher trajectories is highly student-specific and depends critically on their suitability for the student model. Pairing strong teachers (Deepseek-R1) with much weaker students (Qwen-2.5-3B) often fails to yield strong performance, while weaker teachers (Qwen-3-4B-Thinking) can likewise be ineffective when training stronger students (Qwen-3-14B). Moreover, teachers from distant model families (GPT-OSS) often lead to inferior results, suggesting that unfamiliar reasoning patterns are harder for students to absorb. 
Overall, we find no simple teacher-student pairing rule based on surface attributes such as parameter scale or model family, indicating that reasoning data suitability is a nuanced property requiring deeper investigation.

\section{Measuring Data-Student Suitability}
\label{sec:metric}

In this section, we explore metrics for measuring data-student suitability, with the goal of jointly capturing informativeness and alignment.
We begin by introducing two fundamental token-level measures: surprisal and rank (\S~\ref{subsec:metric_pre}), and analyzing the limitations of existing probability-based metrics (\S~\ref{subsec:metric_lim}). 
We then abstract our insights on suitable reasoning data and conduct simulation studies to identify quantitative patterns (\S~\ref{subsec:metric_hyp}). 
Finally, we propose our trajectory-level metric (\S~\ref{subsec:metric_rsr}).

\begin{table*}[t!]
\centering
\resizebox{\linewidth}{!}{%
\begin{tabular}{lcccccc}
\toprule
\multirow{2}{*}{\textbf{Teacher Models}} &
\multirow{2}{*}{\textbf{\makecell{Student\\Performance $\uparrow$}}} &
\multicolumn{2}{c}{\textbf{Probability-based Metrics $\downarrow$}} &
\multicolumn{3}{c}{\textbf{Rank-Surprisal Metrics $\downarrow$}} \\
\cmidrule(lr){3-4}\cmidrule(lr){5-7}
& & Avg-Surprisal & Avg-Surp$_{\text{local}}$  &
Avg-\RatioShort$_{\text{token}}$ & Avg-\RatioShort$^{\text{filter}}_{\text{token}}$ & \textbf{\RatioShort\ (Ours)} \\
\midrule
Qwen-3-8B &
\RankShade{1}{52.0} &
\RankShade{3}{0.65} &
\RankShade{3}{1.16} &
\RankShade{2}{\sci{2.01}{7}} &
\RankShade{1}{3.15} &
\RankShade{1}{2.89} \\
Qwen-3-30B-Thinking &
\RankShade{2}{50.0} &
\RankShade{4}{0.77} &
\RankShade{4}{1.27} &
\RankShade{3}{\sci{2.25}{7}} &
\RankShade{3}{3.47} &
\RankShade{2}{2.95} \\
Nemotron-Super &
\RankShade{3}{48.3} &
\RankShade{2}{0.60} &
\RankShade{2}{1.04} &
\RankShade{6}{\sci{5.51}{7}} &
\RankShade{4}{3.95} &
\RankShade{4}{3.08} \\
Deepseek-R1 &
\RankShade{4}{47.8} &
\RankShade{5}{0.83} &
\RankShade{5}{1.35} &
\RankShade{4}{\sci{2.98}{7}} &
\RankShade{2}{3.31} &
\RankShade{3}{3.00} \\
Magistral-Small &
\RankShade{5}{47.6} &
\RankShade{1}{0.55} &
\RankShade{1}{1.03} &
\RankShade{5}{\sci{3.58}{7}} &
\RankShade{5}{4.38} &
\RankShade{5}{3.09} \\
GPT-OSS-20B &
\RankShade{6}{42.7} &
\RankShade{6}{1.36} &
\RankShade{6}{1.78} &
\RankShade{1}{\sci{5.15}{6}} &
\RankShade{6}{11.10} &
\RankShade{6}{3.83} \\
\bottomrule
\end{tabular}%
}
\caption{Comparison of the student's post-training performance and data-student suitability metrics across trajectories from different teacher models, evaluated on Qwen-2.5-7B.
Darker shading indicates higher performance or better suitability. 
Metrics whose trends align with performance (e.g., \RatioShort) provide more reliable suitability estimates. 
Complete metric scores for all teacher and student models are provided in \S~\ref{app:complete}.
}
\label{tab:rank_shaded_metrics}
\vspace{-1em}
\end{table*}

\vspace{-0.5em}
\subsection{Surprisal and Rank}
\label{subsec:metric_pre}
We first introduce two fundamental methods for quantifying the amount of information a token carries with respect to a student model.
They serve as the building blocks of our metric.

\paragraph{Surprisal (Negative Log-Likelihood)} A common measure is based on the probability of generating the current token $t_k$ given its preceding context $\mathbf{c}_k = (t_1, \ldots, t_{k-1})$ under the student model $\theta$. For numerical stability, probabilities are typically transformed into log space, and the negative log-likelihood—also known as \emph{surprisal}—is used as a measure of informativeness \cite{DBLP:conf/naacl/Hale01}.

\begin{equation}
\label{eq:token_surprisal}
\mathrm{Surprisal}(t_k)
=
-\log p_\theta \bigl(t_k \mid \mathbf{c}_k \bigr)
\end{equation}

\paragraph{Rank} Another method considers the rank of the current token within the model’s prediction distribution over the vocabulary $\mathcal{V}$. Formally, given the conditional distribution $p_\theta(\cdot \mid \mathbf{c}_k)$, the rank of token $t_k$ is defined as the number of tokens with strictly higher probability \cite{DBLP:conf/naacl/RavichanderFSLALMBC25}. 

\begin{equation}
\label{eq:token_rank}
\mathrm{Rank}(t_k)
=
1
+
\sum_{t' \in \mathcal{V}}
\mathbb{I}
\!\left[
p_\theta(t' \mid \mathbf{c}_k)
>
p_\theta(t_k \mid \mathbf{c}_k)
\right]
\end{equation}

Unlike surprisal, rank captures relative familiarity of the token by comparing target token against alternative candidates, revealing signals overlooked by probability-based measures. 
For instance, a token may be assigned a low absolute probability while still ranking among the top candidates.

\subsection{Limitations of Probability-Based Metrics}
\label{subsec:metric_lim}

Existing work primarily relies on log-probability or surprisal to assess data suitability. 
For example, Zhang et al. \cite{DBLP:journals/corr/abs-2502-04194} select trajectories based on the average log-probability of response tokens under the student model. 
Since surprisal is the negation of log-probability, we implement this metric as the average surprisal, denoted as \emph{Avg-Surprisal}.
More recently, Just et al. \cite{DBLP:journals/corr/abs-2510-03988} compute token-level log-probability based on a local context $\mathbf{c}_k^{\text{local}}$ (several preceding sentences), which we implement as average local surprisal (\emph{Avg-Surp\textsubscript{local}}, \S~\ref{app:others}). 
Under these metrics, trajectories with lower surprisal are considered more suitable for the student model.

However, as shown in Table~\ref{tab:rank_shaded_metrics}, lower surprisal (i.e., higher probability) does not necessarily lead to better post-training reasoning performance. 
Both Avg-Surprisal and Avg-Surp\textsubscript{local} assign lower surprisal to trajectories from Nemotron-Super and Magistral-Small, yet training on such trajectories fails to improve reasoning performance.
Similar patterns are observed across all student models, indicating that probability-based metrics tend to favor data that are familiar but insufficiently informative.

At the other extreme, trajectories with very high surprisal (e.g., GPT-OSS-20B) also perform poorly. In contrast, trajectories with moderate surprisal values (e.g., Qwen-3-8B) achieve better results. 
This observation motivates us to investigate the mechanism underlying the surprisal trade-off.

\subsection{Insight and Simulation}
\label{subsec:metric_hyp}
The above analysis suggests that suitable (i.e., effective) teacher trajectories should strike a balance between data informativeness and alignment with the student’s current behavior: they should be neither overly similar to the student’s own generations nor excessively deviant from its prediction distribution.

At first glance, this balance may appear to pose a dilemma. However, we argue that it can be resolved by viewing informativeness and alignment through the lens of \emph{absolute unfamiliarity and relative familiarity}. Informativeness does not require trajectories to be entirely unfamiliar; rather, it suffices that they deviate from the dominant patterns and thus have \textbf{low absolute probability} of being generated by the student. Conversely, alignment does not entail exact agreement with the student’s outputs, but instead requires that the corresponding tokens have \textbf{relatively higher likelihoods than other candidates} in the vocabulary.

Building on this insight, we propose that \emph{effective reasoning trajectories should deviate from the student’s own generations while remaining compatible with the student’s overall generation patterns learned from prior experience}. 
Therefore, tokens in such trajectories are assigned low absolute probability (i.e., high surprisal) by the student model while still ranking relatively high (i.e., having low rank values) in its prediction distribution.

To validate these quantitative patterns and identify features that characterize effective learning trajectories, we conduct a simulation study.

\paragraph{Simulation Setting}
We simulate the student model’s token-level prediction distribution for reasoning trajectories and examine the numerical patterns exhibited by teacher trajectories that we deem effective.
Specifically, we model the student’s prediction distribution as a \emph{bimodal distribution} over the vocabulary $\mathcal{V}$.
The first (major) mode, denoted as $Z_A$, represents tokens that follow the student’s dominant generation patterns, which arise from its general training data.
The second (minor) mode, $Z_B$, occupies a small fraction of the probability mass. It represents tokens from specific generation patterns that deviate from the major mode yet remain familiar to the student model.
We instantiate both $Z_A$ and $Z_B$ as Zipf distributions \cite{zipf2013psycho, mikhaylovskiy2025zipf} over $\mathcal{V}$.
Then the student’s overall token-level prediction distribution $Z$ is constructed as a mixture of $Z_A$ and $Z_B$:
\begin{equation}
\label{eq:sim_student_dist}
\begin{aligned}
& Z = \pi\, Z_A + (1 - \pi)\, Z_B, \\
& Z_A, Z_B \sim \mathrm{Zipf}(\alpha),
\end{aligned}
\end{equation}
where $\pi = \frac{M_A}{M_A + M_B}$ and $M_A > M_B$. 

Based on $Z$, we simulate four types of reasoning trajectories and examine their average token surprisal and rank:
(i) $X_A$, sampled from $Z_A$, representing trajectories that closely follow the student’s dominant generation patterns;
(ii) $X_B$, sampled from $Z_B$, representing trajectories that deviate from the dominant mode while aligning with certain minor patterns within the student;
(iii) $X_C$, sampled from a distribution distinct from both $Z_A$ and $Z_B$, representing misaligned trajectories; and
(iv) $X_D$, sampled from $Z$, representing trajectories that reflect the student’s overall predictive behavior.
More simulation details are provided in \S~\ref{app:sim}.

\begin{figure}[t]
\centering

\begin{minipage}{0.48\linewidth}
\centering
    \includegraphics[width=\linewidth]{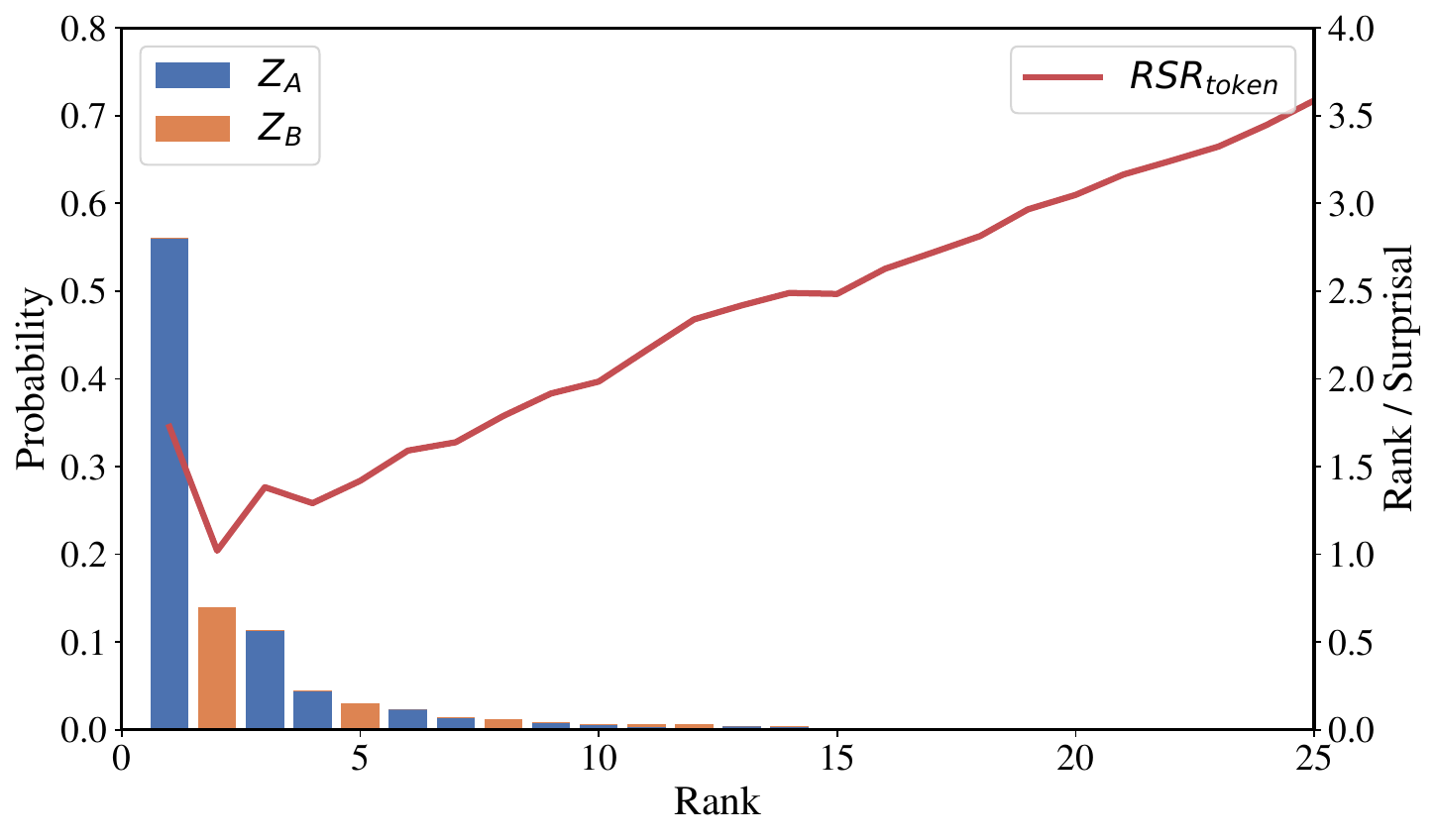}
    \captionof{figure}{Simulation of the student model’s token-level bimodal prediction distribution $Z$. Tokens with low probability yet ranked near the top (typical of the minor mode $Z_B$) tend to yield smaller, and thus preferred, rank-surprisal ratios.}
\label{fig:sim}
\end{minipage}\hfill
\begin{minipage}{0.48\linewidth}
\centering
    \resizebox{\linewidth}{!}{%
    \begin{tabular}{lcccc}
    \toprule
    \textbf{Trajectories} & \textbf{Prob.} & \textbf{Surp.} & \textbf{Rank} & \textbf{\RatioShort$_{\text{token}}$} \\
    \midrule
    $X_A$ (from $Z_A$, major) & 0.41 & 1.38 & 2.49 & 1.69 \\
    \rowcolor{RankColorBase!10}
    $X_B$ (from $Z_B$, minor) & 0.10 & 2.73 & 4.31 & \textbf{1.30} \\
    $X_C$ (from $Z_C$, misaligned) & 0.08 & 4.73 & 11.57 & 2.23 \\
    $X_D$ (from $Z$, mixed)   & 0.35 & 1.67 & 2.93 & 1.62 \\
    \bottomrule
    \end{tabular}%
    }
    \captionof{table}{Simulation results for different types of trajectories, reporting average token probability, surprisal, rank, and \RatioShort$_{\text{token}}$. Effective reasoning trajectories are expected to resemble $X_B$, indicated by the shaded row. \RatioShort$_{\text{token}}$ shows promise as a reliable metric for identifying effective reasoning trajectories (\S~\ref{subsec:metric_rsr}).
    }
\label{tab:simulation_results}
\end{minipage}
\vspace{-1em}
\end{figure}

\paragraph{Simulation Results}
Figure~\ref{fig:sim} presents the simulated bimodal distribution $Z$, where effective reasoning trajectories are expected to align with the distribution of $Z_B$, and thus resemble $X_B$.
As shown in Table~\ref{tab:simulation_results}, $X_B$ exhibits a much higher average surprisal (lower probability) than $X_A$ and $X_D$, while maintaining relatively low rank values (top-ranked), consistent with the quantitative patterns implied by our prior analysis.
Crucially, the results suggest a promising direction for measuring data suitability, which we formalize in \S~\ref{subsec:metric_rsr}.

\subsection{Proposed Metric: \Ratio}
\label{subsec:metric_rsr}

\paragraph{Token-Level Metric}
Motivated by the complementary properties of surprisal and rank, we explore metrics that combine these two signals to jointly capture informativeness and alignment.
One promising choice is the token-level ratio of rank to surprisal, which captures the relative relationship between the two signals and is denoted as \RatioShort$_{\text{token}}$:
\begin{equation}
\label{eq:token_ratio}
\mathrm{\RatioShort}_{\text{token}}(t_k)
\;=\;
\frac{\mathrm{Rank}(t_k)}{\mathrm{Surprisal}(t_k)}.
\end{equation}
Table~\ref{tab:simulation_results} shows that trajectories of the preferred type $X_B$ achieve the lowest average \RatioShort$_{\text{token}}$ (1.30), whereas the misaligned $X_C$ achieve the highest, suggesting that this ratio may be a reliable indicator for identifying effective reasoning trajectories that balance informativeness and alignment.

\paragraph{From Token-Level to Trajectory-Level}
While \RatioShort$_{\text{token}}$ shows encouraging behavior in simulation, directly applying this token-level ratio to assess the overall suitability of a real trajectory presents non-trivial challenges.
In particular, naively averaging \RatioShort$_{\text{token}}$ over all response tokens often leads to large and unstable values (Table~\ref{tab:rank_shaded_metrics}).
This instability stems from tokens that receive extremely high probabilities under certain contexts, yielding near-zero surprisal and unbounded token-level ratios due to division by near-zero values, which dominate the trajectory-level average.

A natural solution is to exclude tokens with very low surprisal when computing the average.
Let $\mathcal{T}_H(\mathbf{x})$ denote the set of tokens whose surprisal lies in the top $H\%$ within a trajectory $\mathbf{x}$.
We define a filtered average as
\begin{equation}
\label{eq:filtered_ratio}
\mathrm{Avg\text{-}\RatioShort}^{\text{filter}}_{\text{token}}(\mathbf{x})
\;=\;
\frac{\sum_{t_k \in \mathcal{T}_H(\mathbf{x})}
\mathrm{\RatioShort}_{\text{token}}(t_k)}{|\mathcal{T}_H(\mathbf{x})|}
\end{equation}
Empirically, we find that using the top 30\% highest-surprisal tokens yields stronger correlation with post-training performance (Table~\ref{tab:rank_shaded_metrics}). 
This suggests that response tokens with higher surprisal have a greater impact on student learning and should therefore be emphasized when computing the average.

Accordingly, instead of hard filtering, we adopt a surprisal-weighted average of the token-level ratios.
A simple derivation shows that this weighted average is equivalent to a trajectory-level ratio between the sum of token ranks and the sum of token surprisals.
For brevity, we denote $r_k = \mathrm{Rank}(t_k)$ and $s_k = \mathrm{Surprisal}(t_k)$:
\begin{equation}
\label{eq:traj_ratio_raw}
\frac{\sum_k s_k \,\mathrm{\RatioShort}_{\text{token}}(t_k)}{\sum_k s_k}
=
\frac{\sum_k s_k \,\frac{r_k}{s_k}}{\sum_k s_k} \\
=
\frac{\sum_k \mathrm{Rank}(t_k)}{\sum_k \mathrm{Surprisal}(t_k)}
\end{equation}
The resulting metric yields a concise form, interpretable as the ratio of the average token rank to the average token surprisal over a trajectory’s response tokens.

In practice, we further observe that extremely unfamiliar tokens can attain very large rank values due to the large vocabulary size, which also leads to numerical instability.
Since such tokens are effectively indistinguishable to the student, we clip rank values at a threshold $r_{max}$ to improve stability without sacrificing meaningful information.
Thus, we define our final trajectory-level metric, \Ratio\ (\RatioShort)\footnote{By default, \RatioShort\ refers to the trajectory-level \RatioShort\ in this paper, unless otherwise specified (e.g., in correlation analysis).}, as
\begin{equation}
\label{eq:traj_ratio}
\mathrm{\RatioShort}(\mathbf{x})
\;=\;
\frac{\sum_{k} \min\!\bigl(\mathrm{Rank}(t_k), r_{max}\bigr)}
{\sum_{k} \mathrm{Surprisal}(t_k)}
\end{equation}

\vspace{-0.2em}
\paragraph{Interpretation}
Our metric admits a simple interpretation.
The numerator, \emph{Rank}, captures relative familiarity: lower rank values indicate that tokens in this trajectory are preferred among alternative candidates by the student and align with the model’s existing behavior.
The denominator, \emph{Surprisal}, captures absolute unfamiliarity: higher surprisal indicates greater deviation from dominant patterns and provides more informative learning signals.
A lower \emph{\RatioShort} therefore identifies trajectories that better balance alignment and informativeness, corresponding to effective reasoning supervision.
\S~\ref{app:case} presents illustrative examples of \RatioShort\ measurements.

\vspace{-0.2em}
\section{Correlation Analysis}
\label{sec:corr}
\vspace{-0.2em}
Preliminary results in Table~\ref{tab:rank_shaded_metrics} have shown that \Ratio\ aligns well with post-training reasoning performance.
To provide a more rigorous evaluation and further demonstrate the effectiveness of \RatioShort\ in measuring data-student suitability, we conduct comprehensive correlation analyses.

\subsection{Main Analysis}
\label{subsec:corr_main}

\begin{table*}[t!]
\centering
\resizebox{\linewidth}{!}{%
\begin{tabular}{llcccccc}
\toprule
\multicolumn{2}{c}{\multirow{2}{*}{\textbf{Metrics}}} &
\multicolumn{6}{c}{\textbf{(Absolute) Spearman Correlation with Post-Training Performance}} \\
\cmidrule(lr){3-8}
& & \textbf{Qwen-3-14B} & \textbf{LLaMA-3.1-8B} & \textbf{Qwen-2.5-7B} & \textbf{Qwen-3-4B} & \textbf{Qwen-2.5-3B} & \textbf{Average}\\
\midrule
\multirow{5}{*}{\makecell[l]{\emph{Student-}\\\emph{Agnostic}}}
& Teacher Params
& 0.04 & 0.34 & 0.2 & 0.02 & 0.26 & 0.01 \\
& Teacher Performance
& 0.49 & 0.34 & 0.13 & 0.23 & 0.03 & 0.23 \\
& Avg-Token Length
& 0.49 & 0.68 & 0.45 & 0.57 & 0.47 & 0.53 \\
& Verified Accuracy
& 0.54 & 0.43 & 0.25 & 0.35 & 0.10 & 0.33 \\
& LLM-judged Quality
& 0.61 & 0.52 & 0.46 & 0.61 & 0.40 & 0.52 \\
& Rule-based Quality
& 0.55 & 0.56 & 0.75 & 0.65 & 0.75 & 0.65 \\
\midrule
\multirow{6}{*}{\makecell[l]{\emph{Student-}\\\emph{Specific}}}
& Avg-Surprisal
& 0.24 & 0.42 & 0.55 & 0.55 & 0.70 & 0.49 \\
& Avg-Surp$_{\text{local}}$
& 0.31 & 0.40 & 0.54 & 0.59 & 0.72 & 0.51 \\
& Avg-Rank
& 0.41 & 0.64 & 0.68 & 0.61 & 0.62 & 0.59 \\
& Influence Score
& 0.52 & 0.19 & 0.32 & 0.47 & 0.59 & 0.11 \\
& G-Norm
& 0.44 & 0.54 & 0.51 & 0.57 & 0.70 & 0.55 \\
& GRACE
& 0.25 & 0.58 & 0.66 & 0.75 & 0.69 & 0.59 \\
& \textbf{\emph{\Ratio}}
& \textbf{0.85} & \textbf{0.85} & \textbf{0.92} & \textbf{0.82} & \textbf{0.85} & \textbf{0.86} \\

\bottomrule
\end{tabular}%
}
\caption{%
Spearman correlation between each data-student suitability metric and post-training reasoning performance (average accuracy on reasoning benchmarks), reported in absolute values for different student models. “Student-agnostic” metrics are computed independently of the specific student model.
}
\label{tab:spearman_metrics}
\vspace{-1em}
\end{table*}

For each of the five student models and each metric, we aggregate trajectory-level suitability (or quality) scores to obtain dataset-level scores for reasoning datasets generated by eleven teacher models (\S~\ref{subsec:exp_setting}).
We then measure the correlation between these dataset-level scores and the student’s reasoning performance after training on the corresponding teacher-generated datasets.
We primarily report Spearman's correlation coefficient, while Pearson correlation exhibits similar trends (\S~\ref{app:res_cor}). 
For dataset-level \RatioShort, we adopt a weighted averaging scheme (\S~\ref{app:dataset_rsr}) similar to Eq.~\ref{eq:traj_ratio_raw}, which yields slightly higher correlation than a simple average of trajectory-level \RatioShort.
We use a clipping threshold of $r_{max}=100$ for \RatioShort\ in all subsequent experiments.
Additional analysis details are provided in \S~\ref{app:corr}, and complete metric scores are reported in \S~\ref{app:complete}.

\paragraph{Compared Metrics} 
We compare \RatioShort\ against a diverse set of metrics for evaluating reasoning trajectories.
These include previously discussed teacher-side indicators (e.g., teacher model performance), basic statistics such as token length, as well as probability-based metrics (e.g., average surprisal and local surprisal \cite{DBLP:journals/corr/abs-2510-03988}) and rank-based metrics.
We also consider commonly used trajectory quality measures, such as rule-based quality scores derived from word frequency \cite{DBLP:journals/corr/abs-2502-03387}, LLM-judged quality scores, and answer accuracy on verifiable questions.
In addition, we include other recent student-specific data suitability metrics, including gradient-based scores (G-Norm and GRACE \cite{DBLP:journals/corr/abs-2511-02833}) and influence scores \cite{DBLP:journals/corr/abs-2510-06108}.

\paragraph{Results} 
Table~\ref{tab:spearman_metrics} shows that \emph{\RatioShortNew consistently exhibits strong correlation with post-training reasoning performance across all student models}, achieving an average Spearman correlation of 0.86 and outperforming all alternative metrics.
These results indicate the effectiveness and practical value of \RatioShort.
In contrast, surprisal-based and rank-based metrics alone yield substantially weaker correlations (at most 0.59). 
Analysis (e.g., \S~\ref{subsec:metric_lim}) suggests that they tend to emphasize high-likelihood trajectories while insufficiently capturing informativeness. 
This highlights the importance of jointly modeling informativeness and alignment via the rank-surprisal ratio.

\vspace{-0.2em}
\subsection{Ablation Study}
\label{subsec:corr_abl}

\begin{wraptable}{r}{0.5\textwidth}
\centering
\vspace{-1.5em}
\resizebox{\linewidth}{!}{%
\begin{tabular}{l S[table-format=1.3] S[table-format=+1.3]}
\toprule
\textbf{Variants} & \textbf{Avg. Corr.} & \textbf{$\Delta$} \\
\midrule
\textbf{\emph{\Ratio} ($r_{max}=100$)} & 0.856 & { } \\
\quad No rank clipping & 0.700 & -0.156 \\
\quad No weighted avg. (Avg-\RatioShort$_{\text{token}}$) & 0.391 & -0.465 \\
\quad Filtered average (Avg-\RatioShort$^{\text{filter}}_{\text{token}}$) & 0.793 & -0.064 \\
\quad Rank clipping: $r_{max}=50$ & 0.696 & -0.160 \\
\quad Rank clipping: $r_{max}=500$ & 0.822 & -0.034 \\
\quad Reduced sample size (200) & 0.864 & 0.007 \\
\bottomrule
\end{tabular}%
}
\caption{Ablation study for \Ratio. $\Delta$ denotes the change in average correlation.}
\label{tab:ablation}
\vspace{-2em}
\end{wraptable}

The derivation of \Ratio\ involves several design components, as well as a hyperparameter $r_{max}$. 
We conduct an ablation study to examine how these choices affect the correlation strength of dataset-level \RatioShort.

As shown in Table~\ref{tab:ablation}, removing either rank clipping or the surprisal-weighted averaging substantially degrades the correlation, validating the necessity of both components in our metric.
In addition, the “Reduced sample size” setting estimates the dataset-level \RatioShort\ using only 200 trajectories per teacher instead of the full 5,000.
The comparable correlations observed under reduced sample size and alternative hyperparameter settings (e.g., $r_{max}=500$) indicate that \RatioShort\ is robust to both data scarcity and reasonable variations in $r_{max}$. Additional ablation results are in \S~\ref{app:abl}.

\section{Practical Applications}
\label{sec:select}

Given the reliable data-student suitability estimation provided by \Ratio\ and its strong correlation with post-training performance, we further examine its practical value as a data selection criterion in two representative scenarios.

\subsection{Trajectory Selection}
\label{subsec:traj_select}

\begin{table*}[t!]
\centering
\resizebox{\linewidth}{!}{%
\begin{tabular}{lccccccccc}
\toprule
\multirow{2}{*}{\textbf{Selection Methods}} &
\multicolumn{5}{c}{\textbf{Qwen-3-14B}} &
\textbf{L3.1-8B} &
\textbf{Q2.5-7B} &
\textbf{Q3-4B} &
\textbf{Q2.5-3B} \\
\cmidrule(lr){2-6}\cmidrule(lr){7-7}\cmidrule(lr){8-8}\cmidrule(lr){9-9}\cmidrule(lr){10-10}
& AIME24 & AIME25 & AMC23 & MATH500 & \textbf{Avg.}
& \textbf{Avg.} & \textbf{Avg.} & \textbf{Avg.} & \textbf{Avg.} \\
\midrule
Random                 & 59.2 & 46.7 & 86.2 & 88.6 & 70.2 & 22.1 & 45.7 & 53.9 & 27.9 \\
Token Length$_{max}$   & 61.7 & 51.7 & 87.5 & 84.8 & 71.4 & 27.3 & 45.4 & 51.3 & 27.1 \\
Rule-based Quality$_{max}$     
                       & 58.3 & 47.5 & 91.3 & 92.0 & 72.3 & 25.8 & 51.6 & 58.0 & 31.2 \\
LLM-judged Quality$_{max}$     
                       & 60.0 & 49.2 & 90.6 & 93.6 & 73.4 & 25.6 & 51.8 & 59.1 & 32.8 \\
Surprisal$_{min}$      & 62.5 & 50.0 & 88.1 & 92.8 & 73.4 & 23.5 & 46.4 & 53.3 & 28.9 \\
G-Norm$_{min}$         & 59.2 & 50.0 & 89.4 & 92.4 & 72.7 & 26.1 & 49.5 & 59.1 & 30.9 \\

\textbf{\emph{\Ratio}}$_{min}$   & \textbf{67.5} & \textbf{59.2} & \textbf{93.1} & \textbf{94.6} & \textbf{78.6} & \textbf{28.5} & \textbf{53.2} & \textbf{61.4} & \textbf{34.8} \\
\bottomrule
\end{tabular}%
}
\caption{
Trajectory selection results showing post-training reasoning performance of student models trained on datasets selected by different methods (5k samples each).
$max$ and $min$ indicate maximizing or minimizing the corresponding metric. 
Model names are abbreviated as \textbf{Q} for Qwen and \textbf{L} for LLaMA. 
Further results, including complete scores and additional GPQA evaluation, are provided in \S~\ref{app:complete} and \S~\ref{app:traj_sel}.
}
\vspace{-1em}
\label{tab:traj_select}
\end{table*}

\paragraph{Experimental Setting}
The trajectory selection task aims to identify the most effective reasoning trajectory from a set of candidates for a given problem or prompt.
This aligns with the widespread need for data-efficient training in real-world scenarios, particularly when resources are limited.
Specifically, we adopt a 33-to-1 setting, where each candidate pool contains 33 trajectories generated by 11 teacher models (3 per teacher; see \S~\ref{subsec:exp_setting}), and one trajectory is selected according to the selection criterion.
This procedure is repeated for all 5,000 training problems and all student models, yielding student-specific 5,000-trajectory teacher datasets for each selection method.
We then fine-tune student models on the constructed datasets and evaluate their post-training reasoning performance on standard math benchmarks, consistent with previous experiments.
We compare \RatioShort\ against a random selection baseline and several previously discussed metrics.
For metric-based methods, candidate trajectories are scored and selected by either maximizing or minimizing the corresponding trajectory-level score.

\paragraph{Results}
As shown in Table~\ref{tab:traj_select}, datasets selected by \Ratio\ consistently achieve the best post-training reasoning performance among all selection methods across student models, demonstrating the effectiveness of \RatioShort\ in identifying suitable trajectories.
Moreover, the performance achieved by \RatioShort\ is comparable to, and for four students even surpasses, the best performance of any single teacher for each student model (Table~\ref{tab:teacher_student_scores}), which serves as a strong upper bound obtained via brute-force search over teacher datasets.
These results underscore the practical value of \RatioShort\ for selecting effective trajectories prior to training.

Complete scores are reported in \S~\ref{app:complete}. Additional trajectory selection results, including GPQA evaluation, No-Selection baseline, further analyses, and reduced-teacher experiments, are provided in \S~\ref{app:traj_sel}.  

\subsection{Teacher Selection}
\label{subsec:teacher_select}

\begin{table*}[t!]
\centering
\resizebox{\textwidth}{!}{%
\begin{tabular}{l*{11}{c}}
\toprule
\multirow{2}{*}{\textbf{Selection Methods}} &
\multicolumn{2}{c}{\textbf{Qwen-3-14B}} &
\multicolumn{2}{c}{\textbf{LLaMA-3.1-8B}} &
\multicolumn{2}{c}{\textbf{Qwen-2.5-7B}} &
\multicolumn{2}{c}{\textbf{Qwen-3-4B}} &
\multicolumn{2}{c}{\textbf{Qwen-2.5-3B}} &
\multirow{2}{*}{\textbf{Avg.}} \\
\cmidrule(lr){2-3}\cmidrule(lr){4-5}\cmidrule(lr){6-7}\cmidrule(lr){8-9}\cmidrule(lr){10-11}
& \textbf{Top-1} & Top-2
& \textbf{Top-1} & Top-2
& \textbf{Top-1} & Top-2
& \textbf{Top-1} & Top-2
& \textbf{Top-1} & Top-2
& \\
\midrule
Teacher Params$_{max}$       & 77.1 & 71.8 & \textbf{28.1} & 22.0 & 47.8 & 45.0 & 55.8 & 53.4 & 29.6 & 26.4 & 45.7 \\
Token Length$_{max}$         & 71.8 & 77.1 & 22.0 & 28.1 & 45.0 & 47.8 & 53.4 & 55.8 & 26.4 & 29.6 & 45.7 \\
Rule-based Quality$_{max}$   & 72.2 & 77.1 & 23.7 & 28.1 & 48.2 & 47.8 & 56.4 & 55.8 & \textbf{33.0} & 29.6 & 47.2 \\
LLM-judged Quality$_{max}$   & 71.8 & 77.2 & 22.0 & 26.7 & 45.0 & 50.0 & 53.4 & 58.8 & 26.4 & 31.2 & 46.3 \\
Surprisal$_{min}$            & 68.8 & 72.2 & 22.8 & 23.7 & 47.6 & 48.2 & 52.2 & 56.4 & 30.6 & 33.0 & 45.6 \\
GRACE$_{min}$ & 72.2 & 68.8 & 22.8 & 28.1 & 47.6 & 48.3 & 52.2 & 58.8 & 30.6 & 26.4 & 45.6 \\
\textbf{\emph{\Ratio}}$_{min}$ & \textbf{77.2} & 77.1 & 26.7 & 28.1 & \textbf{50.0} & 47.8 & \textbf{58.8} & 55.8 & 31.2 & 30.6 & \textbf{48.3} \\
\midrule
Oracle               & 77.2 & 77.1 & 28.1 & 26.7 & 50.0 & 48.2 & 58.8 & 56.4 & 33.0 & 31.2 & 48.7 \\
\bottomrule
\end{tabular}%
}
\caption{Teacher selection results showing post-training reasoning performance of student models trained on data from teachers selected by different methods.
Top-1 and Top-2 denote the highest- and second-highest-ranked teachers for each student.
"Oracle" corresponds to the ground-truth best and second-best teachers.
}
\label{tab:teacher_select}
\vspace{-1em}
\end{table*}

\paragraph{Experimental Setting}
The teacher selection task aims to identify the most suitable teacher model for generating reasoning trajectories to train a given student model prior to distillation, reflecting a recurring challenge in real-world applications.
To better capture practical constraints, we consider a low-resource setting in which generating full training data for every candidate teacher model is either costly or infeasible.
Instead, we sample a small set of 200 trajectories from each candidate teacher, evaluate them using data-student suitability metrics, and select the teacher model based on the resulting dataset-level average score.
We use 6 diverse teacher models (Deepseek-R1, Qwen-3-235B-Thinking, Nemotron-Super, Qwen-3-30B-Thinking, Magistral-Small, and GPT-OSS-20B) as the candidate pool, avoiding consistently well-performing teachers to ensure a non-trivial selection task.

\paragraph{Results}
As shown in Table~\ref{tab:teacher_select}, both the best and second-best teachers selected by \Ratio\ yield strong post-training reasoning performance, achieving average results close to oracle teachers and outperforming other selection methods.
Notably, as also observed in our ablation study (Table~\ref{tab:ablation}), \RatioShort\ remains effective when using only 200 trajectories per teacher for measurement, highlighting its robustness and practical value for identifying suitable teachers in low-resource settings.

\section{Related Work}

\paragraph{Knowledge Distillation}
Knowledge distillation is a powerful approach for transferring knowledge from large models to smaller ones \cite{DBLP:journals/corr/HintonVD15}, and has been widely used in training LLMs \cite{DBLP:journals/corr/abs-2304-03277}.
Prior work shows that stronger teachers do not necessarily yield better students, often due to capability gaps \cite{DBLP:conf/acl/ZhangLSYG025,DBLP:journals/corr/abs-2506-07712} or off-policy data \cite{DBLP:journals/corr/abs-2510-18874}.
Recent studies address this by better aligning teacher supervision with student behavior, achieving an implicit balance through approaches such as on-policy distillation \cite{Agarwal2023OnPolicyDO}, self-distillation \cite{DBLP:conf/nips/ZelikmanWMG22, DBLP:journals/corr/abs-2601-19897}, integration with reinforcement learning \cite{DBLP:journals/corr/abs-2506-07527, DBLP:journals/corr/abs-2506-17211}, SFT optimization \cite{ye2025analyzing, DBLP:journals/corr/abs-2508-05629}, teaching assistants \cite{mirzadeh2020improved, Ding2025MiCoTABT}, uncertainty-based filtering \cite{DBLP:journals/corr/abs-2502-04194}, and interleaved sampling \cite{DBLP:conf/iclr/XuH0LM0WALP25, Peng2025AdaSwitchAS}.
By contrast, our work explicitly quantifies the trade-off between informativeness and alignment in distillation, enabling a principled identification of effective teacher data for a given student.

\paragraph{SFT with Reasoning Trajectories}
Long CoT trajectories provide strong supervision for improving student models’ reasoning performance via SFT \cite{wei2022chain, DBLP:journals/corr/abs-2505-13718}.
In line with findings in the general domain that high-quality data improves SFT \cite{DBLP:conf/nips/ZhouLX0SMMEYYZG23, Diversity}, many studies focus on constructing high-quality CoT data, either by selecting better prompts \cite{DBLP:journals/corr/abs-2504-11919, DBLP:journals/corr/abs-2505-17266} or by filtering reasoning trajectories \cite{DBLP:journals/corr/abs-2502-03387, DBLP:journals/corr/abs-2505-22148, DBLP:journals/corr/abs-2508-09883, DBLP:journals/corr/abs-2506-18896, DBLP:conf/acl/0009SGZ0SSPK0025, liu2025air}.
Recognizing that optimal reasoning data may vary across students, recent work explores student-specific data selection \cite{DBLP:journals/corr/abs-2511-02833, DBLP:journals/corr/abs-2510-03988, DBLP:journals/corr/abs-2510-06108}, an approach widely used in the general domain \cite{DBLP:conf/naacl/LiZLCC0W0024,DBLP:conf/acl/LiCCHGZ24}.
Our work also studies student-specific data selection, but differs from prior work by measuring data-student suitability from the perspective of informative alignment and by conducting more comprehensive evaluations across a wider range of teacher reasoning models.

\section{Broader Applicability of \RatioShort}
\label{sec:broad}

\paragraph{Beyond CoT Data}
Although our work focuses solely on reasoning tasks, \RatioShort\ is not specifically designed for reasoning trajectories (CoTs) and can be applied to general text data. 
Exploring its behavior and effectiveness beyond reasoning settings remains an interesting direction for future work, and we welcome efforts from the community in this direction.

\paragraph{Subset Selection}
Beyond trajectory and teacher selection, \RatioShort\ may also be applicable to subset selection. Subset selection aims to identify a high-quality subset from a dataset composed of different reasoning problems and their corresponding trajectories. Unlike trajectory selection, which compares multiple trajectories for the same problem, subset selection requires cross-problem comparison of \RatioShort\ values to filter samples across heterogeneous reasoning prompts.
This scenario is particularly relevant when only a single trajectory is available per problem, yet improved data efficiency is still desired.
Although subset selection introduces additional confounding factors due to variation across problems—making it harder to isolate trajectory-student suitability effects—\RatioShort\ may still help identify relatively more effective samples across different reasoning problems. Preliminary experiments on internal data provide suggestive evidence for this possibility.

\section{Conclusion}
In this paper, we study data-student suitability in reasoning distillation and propose \Ratio\ (\RatioShort), a simple metric for identifying suitable reasoning trajectories for a given student.
Motivated by our analysis, \RatioShort\ jointly captures a trajectory's informativeness and alignment with the student’s behavior, favoring trajectories with low absolute probability but relatively high-ranked tokens.
Experiments across diverse teacher-student pairs show that \RatioShort\ strongly correlates with post-training performance and consistently outperforms existing metrics.
We further demonstrate its effectiveness in both trajectory and teacher selection.
Overall, our results highlight informative alignment as a promising direction for reasoning distillation.

\section*{Limitations}
Although our work proposes an effective metric for selecting reasoning trajectories to distill student models, the performance of data selection is inherently constrained by the diversity and quality of candidate trajectories or teacher models. When none of the available teacher trajectories are well suited to a given student, the gains from selection alone may be limited. A promising direction for future work is to use our metric to guide the rewriting or synthesis of reasoning trajectories, rather than selecting from a fixed pool.

In addition, given the simple and interpretable form of the derived \RatioShortNew metric, it is natural to ask whether it can be grounded in deeper theoretical principles. However, we have not yet identified a suitable theoretical framework to characterize \RatioShortNew, which we leave for future investigation.

Finally, due to resource constraints, we focus primarily on mathematical reasoning tasks to conduct extensive controlled studies. Although we perform additional evaluation on GPQA to assess the generalizability of \RatioShortNew, we acknowledge that the current experiments still have limitations.
Specifically, we have not yet applied \RatioShortNew to newly generated trajectories beyond the existing mathematical reasoning data, nor to qualitatively different domains such as code or commonsense reasoning.
Systematic cross-domain studies would be valuable, but they require large-scale trajectory regeneration and retraining, entailing substantial computational cost.
Therefore, we focus on the current setting (following prior work) and leave such systematic extensions for future work.

\section*{Acknowledgements}
The authors wish to thank the anonymous reviewers for their helpful comments. This work was partially funded by National Natural Science Foundation of China (No. 62476061, 62521004, 62576106, 62376061).

\bibliographystyle{plain}
\bibliography{refs}

\clearpage
\appendix

\section{Details of Experiments}
\label{app:exp}

\subsection{Determining Training Problem Set}
\label{app:problemset}
All teacher trajectory datasets are constructed using a fixed problem set to avoid confounding effects from variations in problem composition, enabling a more controlled study. We focus on mathematical reasoning and curate 5,000 math problems from the widely used NuminaMath dataset \cite{li2024numinamath}. Following prior work (Sky-T1) \cite{SkyT1}, we apply preprocessing steps such as difficulty filtering to ensure the problem quality.

Specifically, our training dataset consists of problems drawn from the MATH, AIME/AMC, and Olympiads. Sky-T1 provides a difficulty-labeled version of NuminaMath in which each problem is assigned an integer difficulty score from 0 to 9. Using this scale, we randomly sampled 1,667 problems from the MATH subset with difficulty above three, 1,667 problems from the Olympiads subset with difficulty above eight, and 1,666 problems from the AIME/AMC subset, yielding a balanced training set of 5,000 problems. 

We fix the training set size at 5,000 problems for three reasons. First, prior studies~\cite{DBLP:journals/corr/abs-2502-03387, DBLP:journals/corr/abs-2501-19393} have shown that strong reasoning capabilities can be learned from training sets of around 1,000 problems. Second, high-quality reasoning problems are relatively scarce, making further scaling often unrealistic in practice. Third, our distillation study incurs quadratic computational costs across teacher-student pairs. Considering these factors, we find 5,000 problems to offer a good trade-off between representativeness and computational feasibility.

\subsection{Teacher Trajectory Generation}
\label{app:traj_generate}
For each teacher model, we generate reasoning trajectories for 5,000 problems over three independent runs using vLLM \cite{kwon2023efficient} under a maximum generation budget of 31,000 tokens. 
Each independent run (rollout) produces a dataset of 5,000 problem-trajectory pairs, yielding 11 × 3 = 33 datasets in total.

We adopt the officially recommended chat template and model sampling hyperparameters; for example, Qwen models use temperature=$0.6$, top\_p=$0.95$, top\_k=$20$, and min\_p=$0$. Each problem is appended with the instruction:
"Return your final response within \textbackslash boxed\{\}." 
If a sampled trajectory exceeds the token budget, we resample up to 10 times; if all attempts still exceed the budget, we truncate the final trajectory. Across all teacher models, final truncation rates are below 1\%, which helps preserve the quality of training trajectories even for teachers that tend to produce long outputs.

The resulting training dataset of teacher trajectories is formatted as follows: for each sample, we use the original problem statement as the user prompt and a single corresponding teacher-generated trajectory as the assistant response, with a fixed system prompt applied to all instances:
"Please reason step by step, and put your final answer within \textbackslash boxed\{\}." to each problem.

We release all 33 generated trajectory datasets along with \RatioShort-selected subsets (\S~\ref{subsec:traj_select}) on Hugging Face.\footnote{\url{https://huggingface.co/datasets/Umean/RSR_data}}

\subsection{Teacher Models and Student Models}
\label{app:teachers}

\begin{wraptable}{r}{0.35\textwidth}
\centering
\vspace{-2.0em}
\resizebox{\linewidth}{!}{%
\begin{tabular}{l}
\toprule
\textbf{Teacher Models} \\
\midrule
DeepSeek-R1-0528 \\
Qwen-3-235B-Thinking-2507 \\
GPT-OSS-120B (high) \\
LLaMA-3.3-Nemotron-Super-49B-v1.5 \\
QwQ-32B \\
Qwen-3-30B-Thinking-2507 \\
Magistral-Small-2506 \\
GPT-OSS-20B (high) \\
Phi-4-Reasoning-Plus \\
Qwen-3-8B (thinking) \\
Qwen-3-4B-Thinking-2507 \\
\bottomrule
\end{tabular}%
}
\caption{List of teacher models used in our experiments.}
\label{tab:teacher_models}
\vspace{-3.0em}
\end{wraptable}

As discussed earlier, our experiments adopt a more diverse set of teacher models than prior work. 
Specifically, we consider the teacher variants shown in Table~\ref{tab:teacher_models}. 
This set includes several recently emerged reasoning models, extending beyond the DeepSeek-R1 and QwQ teachers commonly used in prior studies.

These teachers produce trajectories that vary in length, informational content, and style. For example, trajectories generated by GPT-OSS models tend to be concise, whereas those from DeepSeek-R1 are generally more detailed.
Specific trajectory examples from different teachers are available in our released datasets on Hugging Face and in our case study (\S~\ref{app:case}).

For the student models, we select five open-source pretrained base models from the Qwen and LLaMA families: Qwen-3-14B, LLaMA-3.1-8B, Qwen-2.5-7B, Qwen-3-4B, and Qwen-2.5-3B. We use base models instead of chat models as students, following the training practice of recent reasoning models. This choice ensures a clean starting point with greater potential \cite{Qwen3}, allows us to observe more substantial training effects, and avoids stylistic mismatch and overlapping supervision between instruction-tuning data and reasoning trajectories.

\subsection{Benchmark Evaluation}
\label{app:bench}
We use vLLM together with the Math-Verify package to evaluate post-trained models on mathematics benchmarks. 
Our evaluated benchmarks, AIME’25, AIME’24, AMC’23, and MATH500, are widely used and span varying difficulty levels, assessing mathematical reasoning and multi-step problem-solving across diverse domains. 
We additionally conduct evaluation on GPQA-Diamond beyond mathematics, as described in \S~\ref{app:traj_sel_gpqa}.
We adopt the Acc@4 metric as the final score, which averages results over four independent evaluations per problem. Evaluating a single model checkpoint typically takes around half an hour using 8 H200 GPUs.

During inference, we use a temperature of $0.6$, a top\_p of $0.95$, and top\_k set to $-1$. 
The maximum generation length is set to 32,768 tokens, consistent with the maximum sequence length used during fine-tuning. Responses that exceed this limit are truncated. This differs from trajectory generation, where we resample multiple times to avoid truncation, as here we aim to evaluate the model’s reasoning ability under a fixed context-length constraint.
Under this setting, the comparable performance of the Qwen-3-235B and Qwen-3-30B in Table~\ref{tab:teacher_student_scores} may be attributable to truncation effects, which we consider a reasonable outcome given the imposed length limit.

\subsection{Details of Model Fine-Tuning}

\begin{wraptable}{r}{0.45\textwidth}
\centering
\vspace{-1em}
\resizebox{\linewidth}{!}{%
\begin{tabular}{lccc}
\toprule
\textbf{Models} & \textbf{Learning Rate} & \textbf{Batch Size} & \textbf{Epoch} \\
\midrule
Qwen-3-14B     & 2.0E-05 & 64 & 8  \\
LLaMA-3.1-8B   & 2.0E-05 & 64 & 10 \\
Qwen-2.5-7B    & 2.0E-05 & 64 & 10 \\
Qwen-3-4B      & 2.0E-05 & 64 & 10 \\
Qwen-2.5-3B    & 5.0E-05 & 64 & 10 \\
\bottomrule
\end{tabular}%
}
\caption{Training hyperparameters for different student models.}
\label{tab:hyp}
\vspace{-2em}
\end{wraptable}

We perform SFT on reasoning trajectory datasets using the LLaMA-Factory~\cite{zheng2024llamafactory} framework and FlashAttention-2~\cite{dao2023flashattention2}. For different student models, we conduct grid search for best set of hyperparameters. The final setting is reported in the Table~\ref{tab:hyp}. During SFT, all student models use a maximum sequence length of 32,768.

Most experiments are conducted on NVIDIA H200 GPUs. A single SFT experiment using the 7B model takes around 7 hours on 8 H200 GPUs for training.

For trajectory selection experiments in \S~\ref{subsec:traj_select}, we fine-tune each selected dataset with three random seeds and report performance averaged over these runs. 
For distillation experiments in \S~\ref{subsec:exp_setting}, however, each teacher already yields three independent trajectory datasets, so we do not further vary random seeds for fine-tuning.

\subsection{Details of Simulation Study}
\label{app:sim}

\begin{wrapfigure}{r}{0.45\linewidth}
    \centering
    \vspace{-2em}
        \includegraphics[width=\linewidth]{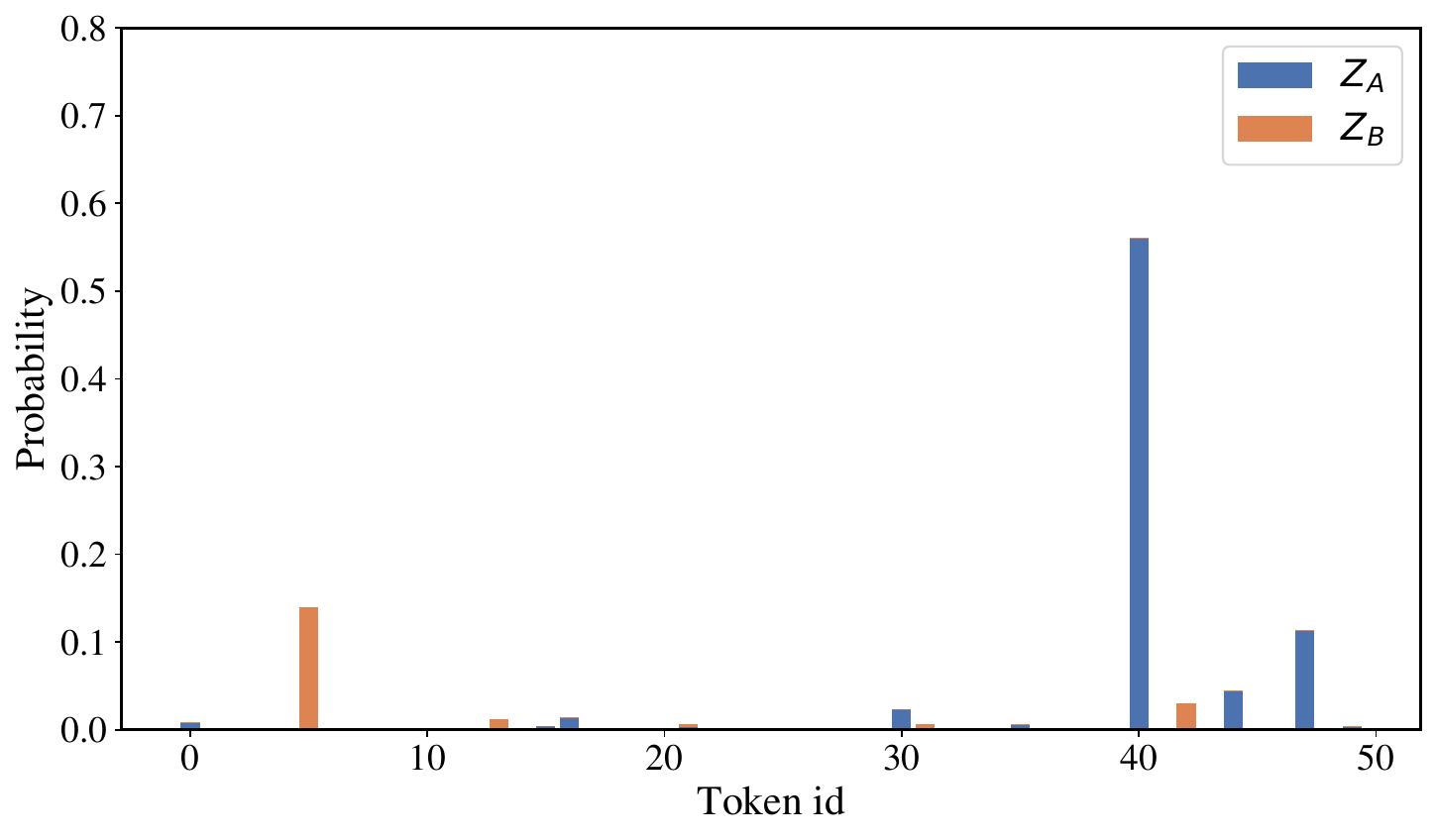}
    \caption{Simulation of the student model’s token-level prediction distribution $Z$, a mixture of Zipf distributions $Z_A$ and $Z_B$.}
    \label{fig:sim_dist}
    \vspace{-3.5em}
\end{wrapfigure}

In practice, we simulate $Z$ by sampling $M_A$ tokens from $Z_A$ and $M_B$ tokens from $Z_B$. 
We use $M_A = 1{,}000{,}000$, $M_B = 250{,}000$, $|\mathcal{V}|=50$, and for each simulated dataset, we sample 10,000 tokens to compute average metrics.
Based on a preliminary fit to reasoning data, we set the Zipf exponent to $\alpha = 2.3$. 

Figure~\ref{fig:sim_dist} depicts the distributions of $Z_A$ and $Z_B$ over the vocabulary, with their mixture representing the simulated token-level prediction distribution $Z$ of the student model. Figure~\ref{fig:sim} is derived from this by ranking the tokens from higher to lower probability.

\subsection{Details of Correlation Analysis}
\label{app:corr}
We use the Spearman correlation coefficient \cite{zar2005spearman} because our analysis focuses on monotonic consistency rather than strict linear relationships between metrics and post-training performance.
We also report Pearson correlation \cite{cohen2009pearson} results in \S~\ref{app:res_cor}. 

We report the absolute values of Spearman correlation coefficients in Table~\ref{tab:spearman_metrics}. For the “Average” values, however, we first average the correlation coefficients across student models and then take the absolute value. We consider this aggregation more appropriate, as correlations with opposite signs across different student models should offset each other; consequently, this averaged value can be lower than the result obtained by averaging absolute correlations.

Since post-training performance is computed by averaging three generation runs per teacher (\S~\ref{subsec:exp_setting}), we likewise average the dataset-level metric over the three trajectory datasets generated by the same teacher to obtain its final score. We observe that our datset-level metric varies marginally across different datasets generated by the same teacher.

\subsection{Definition of Dataset-level \RatioShort}
\label{app:dataset_rsr}

For dataset-level \RatioShort, we adopt a surprisal-weighted averaging scheme analogous to the trajectory-level weighted average (\S~\ref{subsec:metric_rsr}). This design aims to mitigate numerical instability caused by trajectories with disproportionately small average surprisal and to emphasize trajectories with larger average surprisal during dataset-level aggregation.
We now formally define the dataset-level \RatioShort\ used in the correlation analysis (\S~\ref{subsec:corr_main},~\ref{subsec:corr_abl}).

Let $\mathbf{X}=\{\mathbf{x}_j\}_{j=1}^{|\mathbf{X}|}$ denote a dataset of trajectories, where each trajectory $\mathbf{x}_j$ contains response tokens $\{t_{j,k}\}_{k=1}^{|\mathbf{x}_j|}$. For brevity, we define the trajectory-level average clipped rank and average surprisal as
\begin{equation}
\label{eq:traj_avg_rank_surp}
\overline{r}(\mathbf{x}_j)
\;=\;
\frac{1}{|\mathbf{x}_j|}\sum_{k=1}^{|\mathbf{x}_j|} \min\!\bigl(\mathrm{Rank}(t_{j,k}), r_{max}\bigr),
\qquad
\overline{s}(\mathbf{x}_j)
\;=\;
\frac{1}{|\mathbf{x}_j|}\sum_{k=1}^{|\mathbf{x}_j|} \mathrm{Surprisal}(t_{j,k}).
\end{equation}
Accordingly, the trajectory-level \RatioShort\ can be equivalently written as $\mathrm{\RatioShort}(\mathbf{x}_j)=\frac{\overline{r}(\mathbf{x}_j)}{\overline{s}(\mathbf{x}_j)}$. 

To obtain the dataset-level \RatioShort, we apply a weighted averaging scheme analogous to Eq.~\ref{eq:traj_ratio_raw}, using the trajectory-level average surprisal $\overline{s}(\mathbf{x}_j)$ as the weight:
\begin{equation}
\label{eq:dataset_ratio_weighted}
\begin{aligned}
\mathrm{\RatioShort}_{\text{dataset}}(\mathbf{X})
\;&=\;
\frac{\sum_{j=1}^{|\mathbf{X}|}\overline{s}(\mathbf{x}_j)\,\mathrm{\RatioShort}(\mathbf{x}_j)}
{\sum_{j=1}^{|\mathbf{X}|}\overline{s}(\mathbf{x}_j)}\\
\;&=\;
\frac{\sum_{j}\overline{s}(\mathbf{x}_j)\,\frac{\overline{r}(\mathbf{x}_j)}{\overline{s}(\mathbf{x}_j)}}{\sum_{j}\overline{s}(\mathbf{x}_j)}\\
\;&=\;
\frac{\sum_{j}\overline{r}(\mathbf{x}_j)}{\sum_{j}\overline{s}(\mathbf{x}_j)}.
\end{aligned}
\end{equation}
The resulting dataset-level metric admits a concise form, which can be interpreted as the ratio of the summed trajectory-level average clipped ranks to the summed trajectory-level average surprisals.

We compare the correlation strength of the simple dataset-level average of trajectory-level \RatioShort, i.e., $\frac{1}{|\mathbf{X}|}\sum_{j=1}^{|\mathbf{X}|}\mathrm{\RatioShort}(\mathbf{x}_j)$, with that of the dataset-level \RatioShort\ computed via surprisal-weighted averaging (Eq.~\ref{eq:dataset_ratio_weighted}) in the additional ablation study (\S~\ref{app:abl}).

\section{Details of Compared Metrics}

\subsection{LLM-judged Quality}
We use Qwen-3-32B-Instruct in a Non-Thinking configuration as an automatic judge to evaluate reasoning trajectories. Under a fixed evaluation prompt, the judge produces a structured assessment at two levels. First, it assigns dimension-wise scores over five criteria---Factual Accuracy, Logical Rigor, Solution Completeness, Reasoning Efficiency, and Presentation Quality---each accompanied by a brief justification. Second, it aggregates the dimensional assessments into an overall score in $[0,1]$ together with a concise rationale. For dataset-level comparison, we report the mean of the trajectory-level overall scores. The complete evaluation prompt is given in Table~\ref{box:prompt}, adapted from \cite{DBLP:journals/corr/abs-2508-05170}.

\subsection{Rule-based Quality}
We implement the rule-based criterion used for filtering the LIMO Dataset \cite{DBLP:journals/corr/abs-2502-03387}. 
Each response is scored using the following weighted indicators:
\emph{Elaborated reasoning (30\%)}: total word length.
\emph{Self-Verification (20\%)}: frequency of "check" and "verify".
\emph{Exploratory Approach (25\%)}: frequency of "perhaps" and "might".
\emph{Adaptive Granularity (25\%)}: frequency of "therefore" and "since".

To ensure fair comparison across responses of varying lengths, we compute relative keyword frequencies by normalizing absolute keyword counts with respect to the total word count. To account for differences in scale across criteria, we then independently standardize each criterion’s scores into z-scores, which empirically improves correlation.

\subsection{Influence Score}
We adopt the Influence Score method based on the second-order approximation of influence functions~\cite{DBLP:journals/corr/abs-2510-06108}. The core idea is to model an infinitesimal up-weighting of a training sample as a local perturbation to the optimization objective, thereby estimating the resulting marginal change in the evaluation-set loss. Within this framework, a higher influence score indicates that the gradient direction induced by the training sample is more aligned with minimizing the evaluation loss in the vicinity of the converged parameters.

We implement this baseline using the Kronfluence framework. To make the computation tractable for LLMs, we employ the EK-FAC strategy to approximate the Hessian matrix \cite{DBLP:journals/corr/abs-2308-03296}. Specifically, we first precompute the EK-FAC factors on a reference model. Since we do not know which trajectory is better a priori, we use the model trained with randomly selected trajectories per problem as the reference model, corresponding to the “Random” variant in Table~\ref{tab:traj_select}. Then, we compute the pairwise influence scores between the training samples and the evaluation dataset. To optimize for memory and computational efficiency, we apply a low-rank approximation to the query gradients (rank $=4$) and utilize \texttt{bfloat16} precision for the Inverse Hessian-Vector Product calculations. Finally, we average these pairwise scores across the evaluation set to obtain the sample-level (trajectory-level) influence score $s(d)$, which is used as the baseline for evaluating data suitability.

\subsection{G-Norm}
G-Norm \cite{DBLP:journals/corr/abs-2511-02833} assesses the suitability of generated data for a student model by measuring the magnitude of the student’s local gradient signal induced by the data. Formally, G-Norm calculates the magnitude of the loss gradient with respect to the student model's parameters. To address the computational constraints associated with high-dimensional parameter spaces, the method utilizes random projection for dimensionality reduction. For each reasoning trajectory $\mathbf{x}$, we compute the gradient of the loss derived from the student model and apply length normalization to eliminate bias towards longer sequences. These gradients are then projected onto a lower-dimensional subspace via a fixed random matrix, followed by the computation of their $L_2$ norms.

\subsection{GRACE}
We adopt GRACE \cite{DBLP:journals/corr/abs-2511-02833} as a baseline metric for further evaluating the generated reasoning trajectory dataset from a gradient-based perspective. GRACE characterizes the geometry of the optimization landscape by analyzing the spectral structure of the student model gradients. This approach allows for a holistic assessment of how effectively the generated data spans the parameter space required for model optimization.

The calculation proceeds by first projecting the high-dimensional gradients onto a lower-dimensional subspace via a fixed random matrix to ensure computational feasibility. To address the empirical tendency of gradient norms to diminish in longer sequences, the method rescales the projected vectors logarithmically based on the response length. GRACE then computes the metric using a cross-validation strategy where the dataset is divided into multiple partitions. The gradients from a held-out partition are weighted by the regularized inverse covariance matrix estimated from the remaining data. This process effectively quantifies the expected squared norm of the gradients after whitening them with the estimated spectrum of the distribution.

Because GRACE’s cross-validation strategy yields a dataset-level metric rather than a per-sample score, we apply GRACE for teacher selection but not for trajectory-level selection.

\subsection{Others}
\label{app:others}
Here we give formal definitions for Avg-Surprisal and Avg-Surp\textsubscript{local} (\S~\ref{subsec:metric_lim}). 
Given a trajectory $x = (t_1, \ldots, t_T)$,
\begin{equation}
\label{eq:traj_surprisal}
\mathrm{Avg\_Surprisal}(\mathbf{x})
=
\frac{1}{T}
\sum_{k=1}^{T}
-\log p_\theta(t_k \mid \mathbf{c}_k)
\end{equation}
\begin{equation}
\label{eq:traj_local_surprisal}
\mathrm{Avg\_Surp_{local}}(\mathbf{x})
=
\frac{1}{T}
\sum_{k=1}^{T}
-\log p_\theta(t_k \mid \mathbf{c}_k^{\text{local}})
\end{equation}

\section{Computational Cost of \RatioShortNew}
The computation of \Ratio\ requires only a single forward pass through the student model, during which we collect each token’s surprisal and clipped rank from the model logits.
For a single trajectory, computing token-level surprisals and ranks has a worst-case time complexity of $\mathcal{O}(TV)$, where $T$ denotes the number of response tokens and $V$ is the vocabulary size. This computation does not introduce additional complexity beyond that inherent in the dimensionality of the model logits. Moreover, since \RatioShort\ only depends on clipped ranks, the rank computation can be efficiently implemented via top-$k$ selection rather than full-vocabulary comparisons (see our code on GitHub), reducing practical runtime and GPU memory usage.

In practice, computing \RatioShort\ over the 5{,}000-trajectory dataset with a context length of 32{,}768 using a 7B model typically takes under one hour on a single H200 GPU with FlashAttention-2 enabled, which is significantly cheaper than SFT.

We also conduct a direct wall-clock time comparison for computing \RatioShortNew, G-Norm, Influence Score, and LLM-judged Quality on a fixed dataset of 5,000 DeepSeek-R1 trajectories using the LLaMA-3.1-8B student model. As shown in Table~\ref{tab:comp_cost}, \RatioShortNew incurs significantly lower total GPU-hours than the other metrics. This likely stems from the fact that \RatioShortNew requires only a single forward pass through the student model, whereas the other methods involve gradient computations or text generation.

\begin{table*}[ht]
\centering
\resizebox{0.5\linewidth}{!}{%
\begin{tabular}{lccc}
\toprule
\textbf{Metrics} & \textbf{\#GPU} & \textbf{Hours} & \textbf{Total GPU-hours} \\
\midrule
\RatioShort & 1 & 1 & 1 \\
G-Norm & 2 & 3 & 6 \\
Influence Score & 8 & 4 & 32 \\
LLM-judged Quality & 8 & 1.5 & 12 \\
\bottomrule
\end{tabular}%
}
\caption{Computational cost comparison of different metrics on a fixed set of 5,000 DeepSeek-R1 trajectories using the LLaMA-3.1-8B student model, evaluated on H200 GPUs.}
\label{tab:comp_cost}
\end{table*}

\section{What Does \RatioShort\ Prefer? Statistical Analysis and Case Study}

\subsection{Statistical Analysis of \RatioShort-Selected Trajectories}

We compare the statistics of the \RatioShort-selected trajectories with the aggregate statistics (mean/max/min) across source teacher datasets. From Table~\ref{tab:select_stats}, the \RatioShort-selected trajectories exhibit moderate token length and verified accuracy. This suggests that \RatioShortNew is not driven solely by length or accuracy, but instead follows its own evaluation criterion.
Additional analysis of datasets selected by \RatioShortNew can be found in \S~\ref{app:select_analysis}.

\begin{table*}[ht]
\centering
\resizebox{0.55\linewidth}{!}{%
\begin{tabular}{lcc}
\toprule
\textbf{Datasets} & \textbf{Avg. Token Length} & \textbf{Verified Accuracy} \\
\midrule
\textbf{\RatioShortNew Selected} & 9845 & 81.19 \\
Mean over Teachers & 8912 & 81.58 \\
Max. & 13145 & 85.9 \\
Min. & 2552 & 77.6 \\
\bottomrule
\end{tabular}%
}
\caption{Statistics of \RatioShort-selected trajectories and source teacher trajectories in terms of token length and verified accuracy. Qwen-2.5-7B is used as the student model. \RatioShortNew follows its own evaluation criterion, independent of length or accuracy.}
\label{tab:select_stats}
\end{table*}

\subsection{Token-Level Case Study}
\label{app:case}

To better understand how \Ratio\ operates in practice and what kinds of trajectories it prefers, we conduct a token-level case study comparing trajectories generated by different teacher models. 

Table~\ref{box:case_study} presents representative text segments from trajectories generated by GPT-OSS-20B, Nemotron-Super-49B-v1.5, and QwQ-32B, together with their token-level rank, surprisal, and \RatioShort\ values as measured by the Qwen-2.5-7B student, highlighting distinct statistical patterns across teacher models. Unlike Eq.~\ref{eq:token_ratio}, we compute token-level \RatioShort\ with rank clipping at $r_{\max}=100$, consistent with trajectory-level \RatioShort, to improve numerical stability.\footnote{We denote this variant as \RatioShort$^*_{\text{token}}$, and all token-level \RatioShort\ values in this section refer to it.}

As shown, tokens in trajectories from GPT-OSS-20B often exhibit excessively high rank values, suggesting that these trajectories are relatively unfamiliar to the student model and misaligned with the student’s current behavior. This may stem from the fast-paced reasoning flow and uncommon phrasing in GPT-OSS-20B trajectories, which make them harder for the student to follow. 
Meanwhile, tokens from Nemotron-Super frequently have very low surprisals, indicating that these trajectories closely resemble the student’s own generations and therefore provide limited informativeness. This may be attributed to the relatively conventional reasoning flow and limited analytical depth of Nemotron-Super’s trajectories.
As a result, tokens from both models tend to receive high token-level \RatioShort\ values, implying less effective supervision for student training.
By comparison, trajectories from QwQ-32B exhibit a moderate level of comprehensibility and depth. Their tokens show relatively lower rank values alongside higher surprisals, striking a better balance between alignment and informativeness. Consequently, they yield lower token-level \RatioShort\ values and are preferred under the \RatioShort\ criterion. These observations are consistent with the distillation results (Table~\ref{tab:teacher_student_scores}) and the subsequent analysis presented in the main body of the paper. 

We also observe that token-level \RatioShort\ may become numerically unstable when extremely low surprisal yields inflated \RatioShort\ values (e.g., >100). This motivates our surprisal-weighted averaging for trajectory-level \RatioShort\ (\S~\ref{subsec:metric_rsr}), which stabilizes the metric and improves its effectiveness.

Additional trajectory examples from 11 teachers are available in our released datasets on Hugging Face.

\begin{promptbox}{Token-level Case Study}
\noindent \textbf{Trajectories from GPT-OSS-20B:} (High Rank Values, High \RatioShort)
\vspace{1em}

\TokEllipsis
\Tok{Check}{153}{7.28}{0.073}{13.70}
\Tok{no}{504}{9.62}{0.096}{10.42}
\Tok{extra}{18}{4.91}{0.273}{3.66}
\Tok{trick}{1414}{10.69}{0.107}{9.35}\Tok{:}{13}{4.50}{0.346}{2.89}
\Tok{cross}{20}{4.78}{0.239}{4.18}
\Tok{product}{1}{0.24}{0.235}{4.26}
\Tok{magnitude}{6}{3.84}{0.641}{1.56}\Tok{?}{27}{5.66}{0.209}{4.78}
\Tok{Not}{23}{5.03}{0.219}{4.57}
\Tok{needed}{1}{0.91}{0.914}{1.09}\Tok{.}{1}{1.59}{1.594}{0.63}
\Tok{Good}{214}{7.97}{0.080}{12.50}\Tok{.}{2}{2.70}{1.352}{0.74}
\Tok{Thus}{10}{4.44}{0.444}{2.25}
\Tok{final}{2}{2.09}{1.047}{0.96}
\Tok{response}{57}{8.00}{0.140}{7.14}
\Tok{box}{24}{5.91}{0.246}{4.07}
\TokEllipsis

\vspace{1.5em}
\noindent \textbf{Trajectories from Nemotron-Super-49B-v1.5:}  (Low Surprisal, High \RatioShort)
\vspace{1em}

\TokEllipsis
\Tok{But}{2}{1.73}{0.863}{1.16}
\Tok{the}{1}{0.51}{0.512}{1.95}
\Tok{problem}{1}{0.04}{0.039}{25.64}
\Tok{doesn}{1}{0.17}{0.166}{6.02}
\Tok{'t}{1}{0.00}{0.000}{>100}
\Tok{mention}{1}{0.22}{0.217}{4.61}
\Tok{the}{1}{0.29}{0.293}{3.41}
\Tok{height}{1}{0.12}{0.117}{8.55}
\Tok{of}{1}{0.13}{0.134}{7.46}
\Tok{the}{1}{0.00}{0.000}{>100}
\Tok{fountain}{1}{0.00}{0.001}{>100}
\Tok{,}{1}{0.21}{0.212}{4.72}
\Tok{so}{1}{0.17}{0.167}{5.99}
\Tok{maybe}{1}{0.65}{0.648}{1.54}
\Tok{it}{2}{1.45}{0.723}{1.38}
\Tok{'s}{1}{0.13}{0.127}{7.87}
\Tok{safe}{4}{3.03}{0.758}{1.32}
\Tok{to}{1}{0.00}{0.000}{>100}
\Tok{assume}{1}{0.01}{0.006}{166.67}
\Tok{that}{2}{1.05}{0.527}{1.90}
\Tok{the}{1}{0.03}{0.027}{37.04}
\Tok{pl}{1}{0.04}{0.036}{27.78}
\Tok{anks}{1}{0.00}{0.000}{>100}
\Tok{are}{1}{0.02}{0.016}{62.50}
\Tok{placed}{2}{1.40}{0.699}{1.43}
\Tok{in}{1}{0.16}{0.164}{6.10}
\Tok{the}{1}{0.07}{0.068}{14.71}
\Tok{plane}{1}{0.02}{0.021}{47.62}
\Tok{of}{1}{0.06}{0.055}{18.18}
\Tok{the}{1}{0.00}{0.001}{>100}
\Tok{circular}{3}{3.73}{1.245}{0.80}
\Tok{base}{1}{0.00}{0.002}{>100}
\TokEllipsis

\vspace{1.5em}
\noindent \textbf{Trajectories from QwQ-32B:} (Low \RatioShort, Thus Preferred)
\vspace{1em}

\TokEllipsis
\Tok{Let}{4}{2.70}{0.676}{1.48}
\Tok{me}{1}{0.35}{0.354}{2.82}
\Tok{think}{1}{1.63}{1.633}{0.61}
\Tok{of}{5}{2.95}{0.591}{1.69}
\Tok{the}{2}{1.34}{0.672}{1.49}
\Tok{data}{2}{1.94}{0.969}{1.03}
\Tok{as}{1}{0.81}{0.812}{1.23}
\Tok{a}{1}{0.73}{0.734}{1.36}
\Tok{list}{4}{2.88}{0.719}{1.39}
\Tok{.}{2}{1.87}{0.934}{1.07}
\Tok{Let}{1}{1.68}{1.680}{0.60}
\Tok{me}{1}{0.64}{0.641}{1.56}
\Tok{try}{2}{2.00}{1.000}{1.00}
\Tok{to}{1}{0.98}{0.984}{1.02}
\Tok{write}{6}{3.23}{0.539}{1.86}
\Tok{out}{2}{1.15}{0.574}{1.74}
\Tok{the}{1}{0.27}{0.270}{3.70}
\Tok{positions}{21}{5.84}{0.278}{3.60}
\Tok{for}{3}{2.42}{0.807}{1.24}
\Tok{\textvisiblespace}{1}{1.05}{1.055}{0.95}
\Tok{1}{1}{0.08}{0.083}{12.05}
\Tok{0}{1}{0.00}{0.002}{>100}
\Tok{students}{2}{5.25}{2.625}{0.38}
\Tok{?}{425}{12.50}{0.125}{8.00}
\Tok{Wait}{6}{3.38}{0.562}{1.78}
\TokEllipsis

\end{promptbox}

{\captionsetup{type=table}
\captionof{table}{
A case study with visualization of token-level rank, surprisal, and \RatioShort\ values (with rank clipping) across trajectories generated by GPT-OSS-20B, Nemotron-Super-49B-v1.5, and QwQ-32B, as measured by the Qwen-2.5-7B student.
Each token is color-coded according to its \RatioShort\ value, where \colorbox{red!40}{darker red} indicates lower values (stronger preference under \RatioShort). 
GPT-OSS-20B exhibits uncommon phrasing and high rank values; Nemotron-Super exhibits conventional reasoning flow and low surprisal; QwQ-32B strikes a more favorable balance between informativeness and alignment, yielding lower token-level \RatioShort\ values and thus being more likely to be selected.
Trajectory-level \RatioShort\ is defined as the surprisal-weighted average of token-level \RatioShort\ values with rank clipping (\S~\ref{subsec:metric_rsr}).}
\label{box:case_study}}

\section{More Results and Analysis}

\subsection{Additional Results for Correlation Analysis}
\label{app:res_cor}

Table~\ref{tab:ratio_spearman_pearson_abs} reports both Spearman and Pearson correlations for \Ratio. The results show that \RatioShort\ also exhibits strong Pearson correlation with post-training performance. The rationale for primarily using Spearman correlation rather than Pearson correlation is discussed in \S~\ref{app:corr}.

\begin{table*}[ht]
\centering
\resizebox{\linewidth}{!}{%
\begin{tabular}{lccccccc}
\toprule
\multirow{2}{*}{\textbf{Metric}}
& \multirow{2}{*}{\makecell{\textbf{Correlation} \\ \textbf{Measure}}}
& \multicolumn{6}{c}{\textbf{Student Models}} \\
\cmidrule(lr){3-8}
& & Qwen-3-14B
& LLaMA-3.1-8B
& Qwen-2.5-7B
& Qwen-3-4B
& Qwen-2.5-3B
& \textbf{Average} \\
\midrule
\multirow{2}{*}{\emph{\Ratio}}
& Spearman & 0.855 & 0.845 & 0.918 & 0.818 & 0.845 & 0.856 \\
& Pearson & 0.654 & 0.880 & 0.805 & 0.819 & 0.811 & 0.794 \\
\bottomrule
\end{tabular}
}
\caption{Comparison of Spearman and Pearson correlation results on \Ratio\ (absolute values).}
\label{tab:ratio_spearman_pearson_abs}
\end{table*}

\subsection{Additional Ablation Study}
\label{app:abl}

We conduct additional ablation studies with more design variants of \RatioShortNew. Results are shown in Table~\ref{tab:ablation_add}.

\begin{table}[ht]
\centering
\resizebox{0.48\linewidth}{!}{%
\begin{tabular}{l S[table-format=1.3] S[table-format=+1.3]}
\toprule
\textbf{Variants} & \textbf{Avg. Corr.} & \textbf{$\Delta$} \\
\midrule
\textbf{\emph{\Ratio}} & 0.856 & { } \\
\quad Fixed student & 0.785 & -0.071 \\
\quad Simple dataset-level average & 0.838 & -0.018 \\
\quad Use Rank$^{1.05}$ & 0.845 & -0.011 \\
\quad Use Rank$^{0.95}$ & 0.787 & -0.069 \\
\quad Use Surprisal$^{1.05}$ & 0.847 & -0.009\\
\quad Use Surprisal$^{0.95}$ & 0.873 & 0.017 \\
Avg-Rank (clipped) & 0.552 & -0.304 \\
Avg-Entropy & 0.495 & -0.361 \\
Surprisal-Weighted Sum of Rank & 0.604 & -0.252\\
Rank-Weighted Sum of Surprisal & 0.584 & -0.272 \\
Rank Minus Surprisal & 0.585 & -0.271 \\
Rank Times Surprisal & 0.538 & -0.318 \\
Rank-Entropy Ratio & 0.764 & -0.092 \\

\bottomrule
\end{tabular}%
}
\caption{Additional ablation study for \Ratio. $\Delta$ denotes the change in average correlation. All ranks are clipped at $r_{max}=100$.}
\label{tab:ablation_add}
\end{table}

The “Fixed student” variant computes \RatioShort\ using a fixed model (Qwen-3-14B) instead of the target student, and the resulting drop in correlation highlights the importance of student-specific estimation. 
The "Simple dataset-level average" variant computes the dataset-level \RatioShort\ by simply averaging trajectory-level scores, as noted in \S~\ref{app:dataset_rsr}.
The slight degradation in correlation indicates that a simple average of trajectory-level \RatioShort\ remains a robust estimator of dataset-level suitability, while the surprisal-weighted averaging scheme improves the reliability of the aggregated metric at both trajectory-level and dataset-level.

The “Avg-Rank (clipped)” variant applies rank clipping with $r_{max}=100$ to Avg-Rank (see Table~\ref{tab:spearman_metrics}); however, it still exhibits a notable performance gap compared with \RatioShortNew. This suggests that rank information alone is insufficient to yield strong correlation, underscoring the importance of jointly leveraging rank and surprisal.
Similarly, other metrics, such as average entropy and weighted-sum combinations, yield substantially lower correlation than \RatioShortNew. Our analysis suggests that these alternatives still tend to emphasize high-likelihood trajectories while insufficiently capturing informativeness.

The “Rank Minus Surprisal” metric is computed by subtracting surprisal from the (clipped) rank. Although it models the relationship between rank and surprisal, this simple subtraction does not align well with post-training performance; similarly, “Rank Times Surprisal” also shows poor alignment.
The “Rank-Entropy Ratio” is defined as the ratio between the average (clipped) token rank and the average token entropy, where entropy is loosely related to surprisal. Notably, it exhibits a smaller performance drop, suggesting that ratio-based formulations better balance alignment and informativeness.

We also experiment with different exponent choices when computing \Ratio\ (with the default setting corresponding to a power of 1 for both rank and surprisal), for example "use Rank$^{1.05}$". The results indicate that our metric is robust to these variations and may yield higher correlation with hyperparameter tuning. Nevertheless, we use the default power-1 setting to keep the formulation simple.

\subsection{Additional Results for Trajectory Selection}
\label{app:traj_sel}

\subsubsection{\RatioShort-based Trajectory Selection Variants and No-Selection Baseline}

Although trajectory selection enables data-efficient training and offers substantial practical value under resource-constrained settings, we are also interested in how it compares with training on all available data without any selection when resources are sufficient. To this end, we construct a “No Selection” baseline. Unlike \RatioShort, which selects 5k trajectories from different teachers, this baseline directly merges 11 trajectory datasets—one per teacher—into a unified training set comprising 55k samples. 
We do not use the full 165k dataset (i.e., all three trajectory datasets per teacher) for two reasons. First, the computational cost would be prohibitively high. Second, training on multiple highly similar trajectory datasets from the same teacher may reduce data diversity and increase the risk of overfitting. We consider the 55k setting to be sufficiently representative and more consistent with practical use cases.

\begin{table*}[ht]
\centering
\resizebox{\linewidth}{!}{%
\begin{tabular}{lcccccc}
\toprule
\multirow{2}{*}{\textbf{Variant Selection Methods}}
& \textbf{Qwen-3-14B} 
& \textbf{LLaMA-3.1-8B} 
& \textbf{Qwen-2.5-7B} 
& \textbf{Qwen-3-4B} 
& \textbf{Qwen-2.5-3B} 
& \multirow{2}{*}{\textbf{Average}} \\
\cmidrule(lr){2-2}\cmidrule(lr){3-3}\cmidrule(lr){4-4}\cmidrule(lr){5-5}\cmidrule(lr){6-6}
& Math Avg. & Math Avg. & Math Avg. & Math Avg. & Math Avg. & \\
\midrule
\emph{\Ratio}$_{min}$ (5k) & \textbf{78.6} & \textbf{28.5} & \textbf{53.2} & \textbf{61.4} & 34.8 & \textbf{51.3}\\
\quad With Correctness Filtering & 77.5 & 27.9 & 52.3 & 60.7 & \textbf{34.9} & 50.7\\
\quad Fewer Candidates per Teacher & 77.6 & 27.8 & 52.6 & 60.9 & 33.8 & 50.5\\
\midrule
No Selection (55k combined) & 72.7 & 41.5 & 54.2 & 58.9 & 35.3 & 52.5\\
\bottomrule
\end{tabular}%
}
\caption{Performance of \RatioShort-based trajectory selection variants and a No-Selection baseline trained on a 55k combined trajectory dataset. "Math Avg." denotes the average performance over AIME’24, AIME’25, AMC’23, and MATH500.}
\label{tab:ablation_traj}
\end{table*}

From Table~\ref{tab:ablation_traj}, we observe that the 5k trajectories selected by \RatioShortNew achieve post-training performance that is comparable to—or even exceeds—the 55k-sample “No Selection” baseline on most student models. An exception is LLaMA-3.1-8B, which may require larger-scale data to develop reasoning ability due to its relatively limited prior exposure to reasoning tasks. 
In contrast, for stronger students (Qwen-3-14B and Qwen-3-4B), the selected 5k trajectories significantly outperform the unfiltered 55k dataset. This suggests that a small amount of high-quality data is sufficient to effectively activate the reasoning capabilities of student models.
Overall, these findings demonstrate that trajectory selection not only improves data efficiency but can also deliver superior performance with substantially fewer resources, further highlighting the practical value of \RatioShort.

Table~\ref{tab:ablation_traj} also presents trajectory selection results for two selection method variants based on \RatioShortNew, offering further insights.

For "With Correctness Filtering", we consider a combined setting that uses both \RatioShort\ and verified correctness for trajectory selection. Specifically, for problems with verifiably correct trajectories, we first discard incorrect ones and then select among the remaining correct trajectories based on \RatioShort. The results show no significant improvement over selecting trajectories solely based on \RatioShort, particularly for larger student models. This suggests that correctness is sometimes less critical than overall data suitability.

For "Fewer Candidates per Teacher", we evaluate an 11-to-1 setting in which each candidate pool contains 11 teacher trajectories (one per teacher), instead of the original 33-to-1 setting (three per teacher). This setting focuses on selecting the best trajectory for each problem across different teachers, rather than across multiple generations from the same teacher. The results are comparable, with a slight performance gap relative to the 33-to-1 setting, indicating that \RatioShort\ remains effective even when each teacher provides only a single trajectory. The results also suggest that while \RatioShort\ captures suitability differences across generations from the same teacher, such differences are less pronounced than those across different teacher models.

\subsubsection{Additional Evaluation on GPQA}
\label{app:traj_sel_gpqa}

To more comprehensively evaluate the impact of different trajectory selection methods on post-trained models’ reasoning capabilities beyond mathematical problems, we conduct additional evaluation on the GPQA-Diamond benchmark \cite{rein2024gpqa}. GPQA-Diamond consists of 198 challenging multiple-choice questions spanning biology, physics, and chemistry.

\begin{table*}[ht]
\centering
\resizebox{0.95\linewidth}{!}{%
\begin{tabular}{lcccccc}
\toprule
\multirow{2}{*}{\textbf{Selection Methods}}
& \textbf{Qwen-3-14B} 
& \textbf{LLaMA-3.1-8B} 
& \textbf{Qwen-2.5-7B} 
& \textbf{Qwen-3-4B} 
& \textbf{Qwen-2.5-3B} 
& \multirow{2}{*}{\textbf{Average}} \\
\cmidrule(lr){2-2}\cmidrule(lr){3-3}\cmidrule(lr){4-4}\cmidrule(lr){5-5}\cmidrule(lr){6-6}
& GPQA & GPQA & GPQA & GPQA & GPQA & \\
\midrule
Random & 48.5 & 26.8 & 35.4 & 43.4 & 22.2 & 35.3 \\
LLM-judged Quality$_{max}$ & 53.5 & 27.3 & 35.4 & \textbf{45.5} & 21.2 & 36.6 \\
\textbf{\emph{\Ratio}}$_{min}$ & \textbf{55.1} & \textbf{31.3} & \textbf{38.9} & \textbf{45.5} & \textbf{31.3} & \textbf{40.4} \\
\bottomrule
\end{tabular}%
}
\caption{Comparison of different trajectory selection methods on the GPQA-Diamond benchmark across student models. We report Acc@4 as the evaluation metric.}
\label{tab:gpqa}
\end{table*}

Table~\ref{tab:gpqa} summarizes the evaluation results of different trajectory selection methods on GPQA-Diamond. Although the results are less stable than those on mathematical benchmarks due to the out-of-domain nature of this evaluation, datasets selected by \RatioShort\ still achieve the best overall post-training performance. This suggests that \RatioShort\ can identify suitable trajectories that consistently improve student models’ general reasoning capabilities, even when training solely on mathematical problems.

Complete tables combining mathematical and GPQA evaluation results can be found in \S~\ref{app:complete}.

\subsubsection{Analysis of Datasets Selected by \RatioShort}
\label{app:select_analysis}

Table~\ref{tab:dataset_comp} shows the data composition of datasets selected by \RatioShort\ across 11 teacher models. The resulting distributions vary across student models, demonstrating the metric’s ability to select different teacher trajectories tailored to different students. QwQ-32B is generally preferred across student models, consistent with its stable performance. For a clearer contrast in data composition among student models, we refer readers to \S~\ref{app:traj_sel_lesst}, where trajectory selection is performed with fewer teachers and consistently strong teachers such as QwQ-32B are removed. We release the \RatioShort-selected datasets for five student models on Hugging Face.

The average \RatioShort\ values of the selected datasets are also reported in Table~\ref{tab:dataset_comp}. These datasets consistently achieve substantially lower \RatioShort\ values than the teacher trajectory datasets (see \S~\ref{app:complete}), validating that our selection procedure effectively identifies trajectories with low \RatioShort\ for each problem.

\begin{table*}[ht]
\centering
\resizebox{\linewidth}{!}{%
\begin{tabular}{lccccccccccccc}
\toprule
\multirow{2}{*}{\textbf{Selected Datasets}} &
\multicolumn{11}{c}{\textbf{Data Composition over Teacher Models}} &
\multicolumn{2}{c}{\textbf{Metrics}} \\
\cmidrule(lr){2-12}
\cmidrule(lr){13-14}
& R1
& Q3-235B
& GPT-120B
& Nemotron
& QwQ
& Q3-30B
& Magistral
& GPT-20B
& Phi-4
& Q3-8B
& Q3-4B 
& \RatioShort
& Length \\
\midrule
\RatioShort$_{min}$ on Q3-14B    & 6.2\% & 4.5\% & 0.1\% & 0.1\% & 67.3\% & 6.7\% & 1.4\% & 0.0\% & 2.9\% & 2.1\% & 8.6\% & 2.57 & 9363 \\
\RatioShort$_{min}$ on L3.1-8B   & 5.5\% & 1.0\% & 0.0\% & 3.7\% & 48.8\% & 3.9\% & 9.1\% & 0.1\% & 0.2\% & 21.3\% & 6.5\% & 2.69 & 9939 \\
\RatioShort$_{min}$ on Q2.5-7B   & 4.1\% & 1.0\% & 0.7\% & 1.7\% & 55.7\% & 5.3\% & 5.9\% & 0.2\% & 0.8\% & 17.6\% & 6.9\% & 2.67 & 9845 \\
\RatioShort$_{min}$ on Q3-4B     & 2.7\% & 2.2\% & 0.0\% & 0.3\% & 76.6\% & 4.1\% & 3.1\% & 0.0\% & 0.6\% & 4.5\% & 6.1\% & 2.56 & 9419 \\
\RatioShort$_{min}$ on Q2.5-3B   & 2.4\% & 0.8\% & 0.0\% & 2.4\% & 45.1\% & 4.0\% & 12.2\% & 0.0\% & 0.2\% & 26.0\% & 6.6\% & 2.73 & 10169 \\
\bottomrule
\end{tabular}%
}
\caption{Data composition and metric statistics of trajectory datasets selected by \RatioShort\ in the trajectory selection experiments (\S~\ref{subsec:traj_select}) for different student models. Model names are abbreviated as Q for Qwen and L for LLaMA.}
\label{tab:dataset_comp}
\end{table*}

\subsubsection{Additional Trajectory Selection Experiments with Fewer Teacher Models}
\label{app:traj_sel_lesst}

To better reflect practical scenarios in which only a few teachers’ trajectories are available and generally suitable teachers may be absent, we conduct additional experiments that select trajectories from a reduced set of teachers. Specifically, we select trajectories from candidates generated by seven teachers: DeepSeek-R1, Qwen-3-235B-Thinking, Nemotron-Super, Qwen-3-30B-Thinking, Magistral-Small, GPT-OSS-20B, and Qwen-3-8B. This teacher set is formed by combining the representative teachers used in Table~\ref{tab:rank_shaded_metrics} and \S~\ref{subsec:teacher_select}.

The results are shown in Table~\ref{tab:traj_sel_lesst}. Datasets selected by \RatioShort\ still achieve superior post-training reasoning performance compared with the baselines, demonstrating the effectiveness of our metric when only a limited number of teachers are available. We also observe that the performance gap narrows when selecting from seven teachers compared with eleven teachers, which is expected and suggests that a larger candidate space enables trajectory selection to more effectively identify high-quality training data.

\begin{table*}[ht]
\centering
\resizebox{0.5\linewidth}{!}{%
\begin{tabular}{lcc}
\toprule
\multirow{2}{*}{\textbf{Selection Methods}}
& \textbf{Qwen-3-14B}
& \textbf{Qwen-2.5-7B} \\
\cmidrule(lr){2-2}\cmidrule(lr){3-3}
& Math Avg. & Math Avg. \\
\midrule
Random & 72.8 & 47.3 \\
LLM-judged Quality$_{max}$  & 74.4 & 48.3 \\
\textbf{\emph{\Ratio}}$_{min}$  & \textbf{76.8} & \textbf{50.0} \\
\bottomrule
\end{tabular}%
}
\caption{Comparison of different trajectory selection methods under a reduced-teacher setting.
"Math Avg." denotes the average over AIME’24, AIME’25, AMC’23, and MATH500.}
\label{tab:traj_sel_lesst}
\end{table*}

Moreover, Table~\ref{tab:dataset_comp_less} presents the data composition of datasets selected by \RatioShort\ from the reduced set of seven teacher models. The distributions differ markedly for Qwen-3-14B and Qwen-2.5-7B: the former student model tends to favor teachers such as DeepSeek-R1 and Qwen-3-30B, whereas the latter student model shows a stronger preference for smaller models such as Qwen-3-8B. These results further demonstrate the effectiveness of \RatioShort\ in selecting teacher trajectories that are well suited to specific student models.

\begin{table*}[ht]
\centering
\resizebox{\linewidth}{!}{%
\begin{tabular}{lccccccc}
\toprule
\multirow{2}{*}{\textbf{Selected Datasets}} &
\multicolumn{7}{c}{\textbf{Data Composition over Teacher Models}} \\
\cmidrule(lr){2-8}
& Deepseek-R1
& Qwen-3-235B
& Nemotron
& Qwen-3-30B
& Magistral
& GPT-OSS-20B
& Qwen-3-8B \\
\midrule
\RatioShort$_{min}$ on Qwen-3-14B   & 27.40\% & 21.00\% & 1.42\% & \textbf{28.84\%} & 4.92\% & 0.24\% & 16.18\% \\
\RatioShort$_{min}$ on Qwen-2.5-7B & 14.52\% & 8.60\%  & 6.94\% & 20.76\% & 10.64\% & 0.14\% & \textbf{38.40\%} \\
\bottomrule
\end{tabular}%
}
\caption{Data composition of trajectory datasets selected by \RatioShort\ under a reduced-teacher setting (see the setting in \S~\ref{app:traj_sel_lesst} and results in Table~\ref{tab:traj_sel_lesst}). }
\label{tab:dataset_comp_less}
\end{table*}

\section{Complete Results Tables}
\label{app:complete}

Table~\ref{tab:full_qwen14}, \ref{tab:full_llama}, \ref{tab:full_qwen7}, \ref{tab:full_qwen4}, and \ref{tab:full_qwen3} present the full metric assessment results across different teacher trajectory datasets on Qwen-3-14B, LLaMA-3.1-8B, Qwen-2.5-7B, Qwen-3-4B, and Qwen-2.5-3B, respectively.

The complete trajectory selection evaluation results underlying Table~\ref{tab:traj_select} are presented in Table~\ref{tab:traj_select_qwen14b}, \ref{tab:traj_select_llama8b}, \ref{tab:traj_select_qwen7b}, \ref{tab:traj_select_qwen4b}, and \ref{tab:traj_select_qwen3b}.

\section{Others}
\subsection{License for Artifacts and Data Consent}
All artifacts used in this paper are publicly available for academic research purposes, including AIME, AMC, MATH500, GPQA-Diamond and NuminaMath.

\subsection{Data Statement}
The training datasets consist solely of mathematics problems and solutions and contain no offensive content or personal information.

\subsection{AI Assistant Usage Statement}
We used ChatGPT for writing refinement and minor coding assistance. AI assistants were not involved in research innovation, and all core contributions were developed solely by the authors.

\clearpage

\begin{table*}[t!]
\centering
\resizebox{0.9\linewidth}{!}{%
\begin{tabular}{lcccccc}
\toprule
\multirow{2}{*}{\textbf{Selection Methods}} &
\multicolumn{6}{c}{\textbf{Qwen-3-14B}} \\
\cmidrule(lr){2-7}
& AIME'24 & AIME'25 & AMC'23 & MATH500 & \textbf{Math Avg.} & GPQA-Diamond \\
\midrule
Random                       & 59.2 & 46.7 & 86.3 & 88.6 & 70.2 & 48.5 \\
Token Length$_{max}$         & 61.7 & 51.7 & 87.5 & 84.8 & 71.4 & --   \\
Rule-based Quality$_{max}$   & 58.3 & 47.5 & 91.3 & 92.0 & 72.3 & --   \\
LLM-judged Quality$_{max}$   & 60.0 & 49.2 & 90.6 & 93.6 & 73.4 & 53.5 \\
Surprisal$_{min}$            & 62.5 & 50.0 & 88.1 & 92.8 & 73.4 & --   \\
G-Norm$_{min}$               & 59.2 & 50.0 & 89.4 & 92.4 & 72.7 & --   \\
\textbf{\emph{\Ratio}$_{min}$}        & 67.5 & 59.2 & 93.1 & 94.6 & \textbf{78.6} & \textbf{55.1} \\
\bottomrule
\end{tabular}%
}
\caption{
Full post-training reasoning evaluation results for trajectory selection on Qwen-3-14B. 
"Math Avg." denotes the average over AIME’24, AIME’25, AMC’23, and MATH500.
}
\label{tab:traj_select_qwen14b}
\end{table*}

\begin{table*}[t!]
\centering
\resizebox{0.9\linewidth}{!}{%
\begin{tabular}{lcccccc}
\toprule
\multirow{2}{*}{\textbf{Selection Methods}} &
\multicolumn{6}{c}{\textbf{LLaMA-3.1-8B}} \\
\cmidrule(lr){2-7}
& AIME'24 & AIME'25 & AMC'23 & MATH500 & \textbf{Math Avg.} & GPQA-Diamond \\
\midrule
Random                       & 2.5 & 5.8 & 29.4 & 50.8 & 22.1 & 26.8 \\
Token Length$_{max}$         & 8.3 & 6.7 & 36.9 & 57.4 & 27.3 & --   \\
Rule-based Quality$_{max}$   & 6.7 & 9.2 & 29.4 & 58.0 & 25.8 & --   \\
LLM-judged Quality$_{max}$   & 1.7 & 4.2 & 38.1 & 58.6 & 25.6 & 27.3 \\
Surprisal$_{min}$            & 5.0 & 6.7 & 25.6 & 56.8 & 23.5 & --   \\
G-Norm$_{min}$               & 5.8 & 4.2 & 36.9 & 57.6 & 26.1 & --   \\
\textbf{\emph{\Ratio}$_{min}$}        & 5.0 & 8.3 & 36.9 & 63.6 & \textbf{28.5} & \textbf{31.3} \\
\bottomrule
\end{tabular}%
}
\caption{
Full post-training reasoning evaluation results for trajectory selection on LLaMA-3.1-8B.
"Math Avg." denotes the average over AIME’24, AIME’25, AMC’23, and MATH500.
}
\label{tab:traj_select_llama8b}
\end{table*}

\begin{table*}[t!]
\centering
\resizebox{0.9\linewidth}{!}{%
\begin{tabular}{lcccccc}
\toprule
\multirow{2}{*}{\textbf{Selection Methods}} &
\multicolumn{6}{c}{\textbf{Qwen-2.5-7B}} \\
\cmidrule(lr){2-7}
& AIME'24 & AIME'25 & AMC'23 & MATH500 & \textbf{Math Avg.} & GPQA-Diamond \\
\midrule
Random                       & 18.3 & 19.2 & 62.5 & 82.8 & 45.7 & 35.4 \\
Token Length$_{max}$         & 22.5 & 21.7 & 61.9 & 75.6 & 45.4 & --   \\
Rule-based Quality$_{max}$   & 24.2 & 25.0 & 72.5 & 84.8 & 51.6 & --   \\
LLM-judged Quality$_{max}$   & 30.0 & 23.3 & 66.9 & 87.0 & 51.8 & 35.4 \\
Surprisal$_{min}$            & 22.5 & 20.8 & 63.1 & 79.2 & 46.4 & --   \\
G-Norm$_{min}$               & 27.5 & 25.0 & 66.3 & 79.2 & 49.5 & --   \\
\textbf{\emph{\Ratio}$_{min}$}        & 29.2 & 25.8 & 71.3 & 86.6 & \textbf{53.2} & \textbf{38.9} \\
\bottomrule
\end{tabular}%
}
\caption{
Full post-training reasoning evaluation results for trajectory selection on Qwen-2.5-7B.
"Math Avg." denotes the average over AIME’24, AIME’25, AMC’23, and MATH500.
}
\label{tab:traj_select_qwen7b}
\end{table*}

\begin{table*}[t!]
\centering
\resizebox{0.9\linewidth}{!}{%
\begin{tabular}{lcccccc}
\toprule
\multirow{2}{*}{\textbf{Selection Methods}} &
\multicolumn{6}{c}{\textbf{Qwen-3-4B}} \\
\cmidrule(lr){2-7}
& AIME'24 & AIME'25 & AMC'23 & MATH500 & \textbf{Math Avg.} & GPQA-Diamond \\
\midrule
Random                       & 30.8 & 30.8 & 68.8 & 85.2 & 53.9 & 43.4 \\
Token Length$_{max}$         & 28.3 & 28.3 & 71.9 & 76.6 & 51.3 & --   \\
Rule-based Quality$_{max}$   & 36.7 & 32.5 & 77.5 & 85.4 & 58.0 & --   \\
LLM-judged Quality$_{max}$   & 37.5 & 32.5 & 77.5 & 88.8 & 59.1 & \textbf{45.5} \\
Surprisal$_{min}$            & 29.2 & 33.3 & 71.9 & 78.8 & 53.3 & --   \\
G-Norm$_{min}$               & 34.2 & 33.3 & 79.4 & 89.4 & 59.1 & --   \\
\textbf{\emph{\Ratio}$_{min}$}        & 44.2 & 35.0 & 77.5 & 88.8 & \textbf{61.4} & \textbf{45.5} \\
\bottomrule
\end{tabular}%
}
\caption{
Full post-training reasoning evaluation results for trajectory selection on Qwen-3-4B.
"Math Avg." denotes the average over AIME’24, AIME’25, AMC’23, and MATH500.
}
\label{tab:traj_select_qwen4b}
\end{table*}

\begin{table*}[t!]
\centering
\resizebox{0.9\linewidth}{!}{%
\begin{tabular}{lcccccc}
\toprule
\multirow{2}{*}{\textbf{Selection Methods}} &
\multicolumn{6}{c}{\textbf{Qwen-2.5-3B}} \\
\cmidrule(lr){2-7}
& AIME'24 & AIME'25 & AMC'23 & MATH500 & \textbf{Math Avg.} & GPQA-Diamond \\
\midrule
Random                       & 6.7 & 7.5 & 36.3 & 61.2 & 27.9 & 22.2 \\
Token Length$_{max}$         & 5.0 & 10.0 & 37.5 & 56.0 & 27.1 & --   \\
Rule-based Quality$_{max}$   & 5.8 & 10.0 & 45.0 & 63.8 & 31.2 & --   \\
LLM-judged Quality$_{max}$   & 12.5 & 10.8 & 39.4 & 68.4 & 32.8 & 21.2 \\
Surprisal$_{min}$            & 5.8 & 8.3 & 39.4 & 62.0 & 28.9 & --   \\
G-Norm$_{min}$               & 9.2 & 9.2 & 46.3 & 59.0 & 30.9 & --   \\
\textbf{\emph{\Ratio}$_{min}$}        & 11.7 & 11.7 & 45.0 & 70.8 & \textbf{34.8} & \textbf{31.3} \\
\bottomrule
\end{tabular}%
}
\caption{
Full post-training reasoning evaluation results for trajectory selection on Qwen-2.5-3B.
"Math Avg." denotes the average over AIME’24, AIME’25, AMC’23, and MATH500.
}
\label{tab:traj_select_qwen3b}
\vspace{2em}
\end{table*}

\begin{table*}[t]
\centering
\resizebox{\textwidth}{!}{%
\begin{tabular}{lrrrrrrrrrrr}
\toprule
\textbf{Metrics} &
\textbf{Deepseek-R1} &
\textbf{Q3-235B} &
\textbf{GPT-120B} &
\textbf{Nemotron} &
\textbf{QwQ-32B} &
\textbf{Q3-30B} &
\textbf{Magistral} &
\textbf{GPT-20B} &
\textbf{Phi-4} &
\textbf{Q3-8B} &
\textbf{Q3-4B} \\
\midrule
Avg-Token Length & 12077.7 & 12571.1 & 2552.3 & 8798.2 & 9070.4 & 10887.1 & 10993.3 & 3822.7 & 3643.1 & 10473.9 & 13145.8 \\
Teacher Performance             & 0.911 & 0.912 & 0.883 & 0.823 & 0.852 & 0.923 & 0.710 & 0.834 & 0.727 & 0.825 & 0.873 \\
Verified Accuracy               & 0.849 & 0.859 & 0.801 & 0.786 & 0.820 & 0.857 & 0.776 & 0.784 & 0.804 & 0.803 & 0.835 \\
Rule-based Quality              & 0.226 & -0.031 & -0.484 & 0.259 & 0.395 & -0.106 & 0.100 & -0.393 & -0.389 & 0.465 & -0.042 \\
LLM-judged Quality              & 0.908 & 0.966 & 0.896 & 0.882 & 0.901 & 0.963 & 0.815 & 0.863 & 0.823 & 0.911 & 0.951 \\
G-Norm                          & 33.615 & 33.581 & 72.106 & 31.801 & 39.701 & 34.569 & 35.455 & 66.665 & 62.448 & 33.913 & 33.580 \\
Influence Score ($\times 10^{5}$) & 0.161 & 0.696 & 0.418 & 1.447 & 0.795 & 0.671 & -0.008 & 0.244 & -0.348 & 0.816 & 0.633 \\
Avg-Surprisal                   & 0.660 & 0.616 & 1.162 & 0.424 & 0.629 & 0.591 & 0.410 & 1.276 & 1.068 & 0.485 & 0.581 \\
Avg-Rank                        & 49.412 & 55.609 & 430.379 & 65.293 & 55.249 & 58.448 & 41.968 & 365.886 & 303.840 & 45.098 & 49.145 \\
Avg-Rank (clipped)              & 1.930 & 1.811 & 4.097 & 1.422 & 1.682 & 1.728 & 1.355 & 4.652 & 3.589 & 1.457 & 1.695 \\
GRACE                           & 0.028 & 0.028 & 0.168 & 0.025 & 0.031 & 0.030 & 0.025 & 0.154 & 0.113 & 0.023 & 0.026 \\
Avg-\RatioShort$_{\text{token}}$ ($\times 10^{8}$) & 1.564 & 2.970 & 0.706 & 2.598 & 0.768 & 4.078 & 31.957 & 0.348 & 0.315 & 2.463 & 2.569 \\
Avg-\RatioShort$^{\text{filter}}_{\text{token}}$ & 8.713 & 9.962 & 62.383 & 13.663 & 9.674 & 10.437 & 22.701 & 53.858 & 43.930 & 8.799 & 9.079 \\
\RatioShort\ (200 sample)                & 2.916 & 2.943 & 3.504 & 3.342 & 2.684 & 2.915 & 3.252 & 3.679 & 3.348 & 2.961 & 2.904 \\
\RatioShort\                             & 2.925 & 2.940 & 3.527 & 3.352 & 2.673 & 2.923 & 3.302 & 3.645 & 3.360 & 3.003 & 2.918 \\
\midrule
Post-Training Performance & 77.1 & 71.8 & 66.7 & 72.2 & 77.4 & 77.2 & 68.8 & 69.5 & 54.1 & 74.6 & 76.8 \\
\bottomrule
\end{tabular}%
}
\caption{Full metric assessment results on Qwen-3-14B. 
Model names are abbreviated with Q for Qwen.}
\label{tab:full_qwen14}
\end{table*}

\begin{table*}[t]
\centering
\resizebox{\textwidth}{!}{%
\begin{tabular}{lrrrrrrrrrrr}
\toprule
\textbf{Metrics} &
\textbf{Deepseek-R1} &
\textbf{Q3-235B} &
\textbf{GPT-120B} &
\textbf{Nemotron} &
\textbf{QwQ-32B} &
\textbf{Q3-30B} &
\textbf{Magistral} &
\textbf{GPT-20B} &
\textbf{Phi-4} &
\textbf{Q3-8B} &
\textbf{Q3-4B} \\
\midrule
Avg-Token Length & 12077.7 & 12571.1 & 2552.3 & 8798.2 & 9070.4 & 10887.1 & 10993.3 & 3822.7 & 3643.1 & 10473.9 & 13145.8 \\
Teacher Performance             & 0.911 & 0.912 & 0.883 & 0.823 & 0.852 & 0.923 & 0.710 & 0.834 & 0.727 & 0.825 & 0.873 \\
Verified Accuracy               & 0.849 & 0.859 & 0.801 & 0.786 & 0.820 & 0.857 & 0.776 & 0.784 & 0.804 & 0.803 & 0.835 \\
Rule-based Quality              & 0.226 & -0.031 & -0.484 & 0.259 & 0.395 & -0.106 & 0.100 & -0.393 & -0.389 & 0.465 & -0.042 \\
LLM-judged Quality              & 0.908 & 0.966 & 0.896 & 0.882 & 0.901 & 0.963 & 0.815 & 0.863 & 0.823 & 0.911 & 0.951 \\
G-Norm                          & 52.768 & 55.501 & 120.459 & 55.882 & 56.478 & 57.108 & 43.666 & 109.119 & 95.909 & 48.365 & 53.058 \\
Influence Score ($\times 10^{6}$) & 2.865 & 1.330 & 2.361 & 1.123 & 1.359 & 1.258 & 0.932 & 2.755 & 2.205 & 1.125 & 1.243 \\
Avg-Surprisal                   & 0.945 & 0.927 & 1.418 & 0.724 & 0.953 & 0.899 & 0.668 & 1.530 & 1.277 & 0.754 & 0.866 \\
Avg-Rank                        & 10.328 & 11.101 & 72.356 & 11.762 & 11.028 & 11.206 & 8.463 & 66.664 & 49.643 & 8.721 & 9.757 \\
Avg-Rank (clipped)              & 2.831 & 2.821 & 5.633 & 2.183 & 2.687 & 2.665 & 2.016 & 6.178 & 4.638 & 2.174 & 2.552 \\
GRACE                           & 0.162 & 0.177 & 1.005 & 0.185 & 0.183 & 0.181 & 0.111 & 0.853 & 0.639 & 0.143 & 0.159 \\
Avg-\RatioShort$_{\text{token}}$ ($\times 10^{8}$) & 1.073 & 0.872 & 0.588 & 0.976 & 0.578 & 0.972 & 1.843 & 0.284 & 0.555 & 0.727 & 0.934 \\
Avg-\RatioShort$^{\text{filter}}_{\text{token}}$ & 3.604 & 3.791 & 18.711 & 4.019 & 3.706 & 3.805 & 7.881 & 17.525 & 13.173 & 3.206 & 3.458 \\
\RatioShort\ (200 sample)                & 2.995 & 3.044 & 3.976 & 2.997 & 2.848 & 2.960 & 3.000 & 4.104 & 3.608 & 2.857 & 2.946 \\
\RatioShort\                             & 2.996 & 3.044 & 3.971 & 3.016 & 2.818 & 2.965 & 3.020 & 4.038 & 3.633 & 2.882 & 2.945 \\
\midrule
Post-Training Performance & 28.1 & 22.0 & 15.2 & 23.7 & 27.1 & 26.7 & 22.8 & 17.9 & 14.5 & 26.5 & 28.2 \\
\bottomrule
\end{tabular}%
}
\caption{Full metric assessment results on LLaMA-3.1-8B.
Model names are abbreviated with Q for Qwen.}
\label{tab:full_llama}
\end{table*}

\begin{table*}[t]
\centering
\resizebox{\textwidth}{!}{%
\begin{tabular}{lrrrrrrrrrrr}
\toprule
\textbf{Metrics} &
\textbf{Deepseek-R1} &
\textbf{Q3-235B} &
\textbf{GPT-120B} &
\textbf{Nemotron} &
\textbf{QwQ-32B} &
\textbf{Q3-30B} &
\textbf{Magistral} &
\textbf{GPT-20B} &
\textbf{Phi-4} &
\textbf{Q3-8B} &
\textbf{Q3-4B} \\
\midrule
Avg-Token Length & 12077.7 & 12571.1 & 2552.3 & 8798.2 & 9070.4 & 10887.1 & 10993.3 & 3822.7 & 3643.1 & 10473.9 & 13145.8 \\
Teacher Performance             & 0.911 & 0.912 & 0.883 & 0.823 & 0.852 & 0.923 & 0.710 & 0.834 & 0.727 & 0.825 & 0.873 \\
Verified Accuracy               & 0.849 & 0.859 & 0.801 & 0.786 & 0.820 & 0.857 & 0.776 & 0.784 & 0.804 & 0.803 & 0.835 \\
Rule-based Quality              & 0.226 & -0.031 & -0.484 & 0.259 & 0.395 & -0.106 & 0.100 & -0.393 & -0.389 & 0.465 & -0.042 \\
LLM-judged Quality              & 0.908 & 0.966 & 0.896 & 0.882 & 0.901 & 0.963 & 0.815 & 0.863 & 0.823 & 0.911 & 0.951 \\
G-Norm                          & 42.327 & 39.127 & 84.024 & 38.592 & 45.587 & 40.408 & 32.527 & 78.025 & 73.562 & 38.400 & 37.860 \\
Influence Score ($\times 10^{6}$) & -0.520 & -0.766 & -0.812 & -0.732 & -0.737 & -0.767 & -0.466 & -0.788 & -0.740 & -0.727 & -0.770 \\
Avg-Surprisal                   & 0.825 & 0.799 & 1.236 & 0.597 & 0.820 & 0.767 & 0.553 & 1.356 & 1.131 & 0.647 & 0.748 \\
Avg-Rank                        & 6.192 & 6.438 & 36.628 & 6.413 & 6.312 & 6.411 & 4.724 & 35.233 & 25.818 & 5.000 & 5.683 \\
Avg-Rank (clipped)              & 2.477 & 2.416 & 4.557 & 1.842 & 2.280 & 2.264 & 1.710 & 5.189 & 3.921 & 1.869 & 2.198 \\
GRACE                           & 0.168 & 0.199 & 1.760 & 0.126 & 0.160 & 0.142 & 0.119 & 1.491 & 0.908 & 0.120 & 0.139 \\
Avg-\RatioShort$_{\text{token}}$ ($\times 10^{7}$) & 2.980 & 2.694 & 1.104 & 5.508 & 3.055 & 2.253 & 3.582 & 0.515 & 0.824 & 2.009 & 2.590 \\
Avg-\RatioShort$^{\text{filter}}_{\text{token}}$ & 2.671 & 2.709 & 9.663 & 3.063 & 2.542 & 2.705 & 4.359 & 9.553 & 7.163 & 2.434 & 2.551 \\
\RatioShort\ (200 sample)                & 2.999 & 3.030 & 3.670 & 3.066 & 2.803 & 2.950 & 3.067 & 3.887 & 3.471 & 2.864 & 2.935 \\
\RatioShort\                             & 3.002 & 3.023 & 3.686 & 3.086 & 2.779 & 2.951 & 3.091 & 3.827 & 3.468 & 2.888 & 2.940 \\
\midrule
Post-Training Performance & 47.3 & 45.0 & 40.7 & 48.3 & 52.0 & 50.0 & 47.6 & 42.7 & 35.2 & 52.0 & 51.8 \\
\bottomrule
\end{tabular}%
}
\caption{Full metric assessment results on Qwen-2.5-7B.
Model names are abbreviated with Q for Qwen.}
\label{tab:full_qwen7}
\end{table*}

\begin{table*}[t]
\centering
\resizebox{\textwidth}{!}{%
\begin{tabular}{lrrrrrrrrrrr}
\toprule
\textbf{Metrics} &
\textbf{Deepseek-R1} &
\textbf{Q3-235B} &
\textbf{GPT-120B} &
\textbf{Nemotron} &
\textbf{QwQ-32B} &
\textbf{Q3-30B} &
\textbf{Magistral} &
\textbf{GPT-20B} &
\textbf{Phi-4} &
\textbf{Q3-8B} &
\textbf{Q3-4B} \\
\midrule
Avg-Token Length & 12077.7 & 12571.1 & 2552.3 & 8798.2 & 9070.4 & 10887.1 & 10993.3 & 3822.7 & 3643.1 & 10473.9 & 13145.8 \\
Teacher Performance             & 0.911 & 0.912 & 0.883 & 0.823 & 0.852 & 0.923 & 0.710 & 0.834 & 0.727 & 0.825 & 0.873 \\
Verified Accuracy               & 0.849 & 0.859 & 0.801 & 0.786 & 0.820 & 0.857 & 0.776 & 0.784 & 0.804 & 0.803 & 0.835 \\
Rule-based Quality              & 0.226 & -0.031 & -0.484 & 0.259 & 0.395 & -0.106 & 0.100 & -0.393 & -0.389 & 0.465 & -0.042 \\
LLM-judged Quality              & 0.908 & 0.966 & 0.896 & 0.882 & 0.901 & 0.963 & 0.815 & 0.863 & 0.823 & 0.911 & 0.951 \\
G-Norm                          & 33.263 & 33.036 & 70.922 & 35.749 & 43.248 & 33.884 & 39.333 & 67.150 & 62.701 & 38.211 & 33.109 \\
Influence Score ($\times 10^{4}$) & -0.284 & -0.013 & -1.077 & 3.601 & 1.919 & 0.068 & -1.214 & -1.202 & -0.965 & 1.395 & -1.149 \\
Avg-Surprisal                   & 0.721 & 0.685 & 1.208 & 0.474 & 0.693 & 0.650 & 0.450 & 1.319 & 1.102 & 0.526 & 0.615 \\
Avg-Rank                        & 9.172 & 10.267 & 82.658 & 10.747 & 9.773 & 10.346 & 7.503 & 72.151 & 54.562 & 7.710 & 9.011 \\
Avg-Rank (clipped)              & 2.132 & 2.023 & 4.581 & 1.526 & 1.835 & 1.896 & 1.422 & 5.130 & 3.955 & 1.535 & 1.802 \\
GRACE                           & 0.148 & 0.151 & 0.524 & 0.148 & 0.165 & 0.145 & 0.139 & 0.436 & 0.425 & 0.124 & 0.116 \\
Avg-\RatioShort$_{\text{token}}$ ($\times 10^{8}$) & 1.314 & 1.785 & 0.361 & 4.915 & 1.122 & 2.187 & 29.884 & 0.203 & 0.182 & 3.123 & 2.229 \\
Avg-\RatioShort$^{\text{filter}}_{\text{token}}$ & 3.641 & 4.058 & 21.345 & 6.773 & 3.619 & 4.197 & 16.064 & 18.829 & 14.369 & 4.010 & 3.912 \\
\RatioShort\ (200 sample)                & 2.947 & 2.961 & 3.771 & 3.197 & 2.660 & 2.908 & 3.124 & 3.937 & 3.579 & 2.881 & 2.919 \\
\RatioShort\                             & 2.958 & 2.955 & 3.794 & 3.216 & 2.649 & 2.917 & 3.160 & 3.888 & 3.588 & 2.919 & 2.928 \\
\midrule
Post-Training Performance & 55.8 & 53.4 & 47.9 & 56.4 & 61.2 & 58.8 & 52.2 & 48.4 & 40.2 & 61.2 & 61.9 \\
\bottomrule
\end{tabular}%
}
\caption{Full metric assessment results on Qwen-3-4B.
Model names are abbreviated with Q for Qwen.}
\label{tab:full_qwen4}
\end{table*}

\begin{table*}[t]
\centering
\resizebox{\textwidth}{!}{%
\begin{tabular}{lrrrrrrrrrrr}
\toprule
\textbf{Metrics} &
\textbf{Deepseek-R1} &
\textbf{Q3-235B} &
\textbf{GPT-120B} &
\textbf{Nemotron} &
\textbf{QwQ-32B} &
\textbf{Q3-30B} &
\textbf{Magistral} &
\textbf{GPT-20B} &
\textbf{Phi-4} &
\textbf{Q3-8B} &
\textbf{Q3-4B} \\
\midrule
Avg-Token Length & 12077.7 & 12571.1 & 2552.3 & 8798.2 & 9070.4 & 10887.1 & 10993.3 & 3822.7 & 3643.1 & 10473.9 & 13145.8 \\
Teacher Performance & 0.911 & 0.912 & 0.883 & 0.823 & 0.852 & 0.923 & 0.710 & 0.834 & 0.727 & 0.825 & 0.873 \\
Verified Accuracy & 0.849 & 0.859 & 0.801 & 0.786 & 0.820 & 0.857 & 0.776 & 0.784 & 0.804 & 0.803 & 0.835 \\
Rule-based Quality & 0.226 & -0.031 & -0.484 & 0.259 & 0.395 & -0.106 & 0.100 & -0.393 & -0.389 & 0.465 & -0.042 \\
LLM-judged Quality & 0.908 & 0.966 & 0.896 & 0.882 & 0.901 & 0.963 & 0.815 & 0.863 & 0.823 & 0.911 & 0.951 \\
G-Norm & 29.027 & 27.225 & 74.196 & 27.733 & 30.065 & 27.843 & 26.490 & 67.096 & 62.386 & 25.996 & 26.518 \\
Influence Score ($\times 10^{5}$) & -1.288 & -2.927 & -2.157 & -2.873 & -3.711 & -2.730 & -1.694 & -2.311 & -2.555 & -3.225 & -2.812 \\
Avg-Surprisal & 0.903 & 0.885 & 1.346 & 0.657 & 0.891 & 0.847 & 0.608 & 1.454 & 1.210 & 0.704 & 0.822 \\
Avg-Rank & 18.247 & 19.034 & 145.459 & 24.048 & 21.422 & 19.906 & 16.483 & 119.646 & 90.940 & 16.967 & 17.978 \\
Avg-Rank (clipped) & 2.794 & 2.747 & 5.307 & 2.040 & 2.550 & 2.551 & 1.855 & 5.870 & 4.381 & 2.049 & 2.462 \\
GRACE & 0.096 & 0.092 & 0.602 & 0.093 & 0.100 & 0.147 & 0.077 & 0.991 & 0.426 & 0.075 & 0.085 \\
Avg-\RatioShort$_{\text{token}}$ ($\times 10^{8}$) & 0.953 & 0.805 & 0.177 & 2.848 & 1.916 & 0.746 & 2.188 & 0.0768 & 1.814 & 0.723 & 0.791 \\
Avg-\RatioShort$^{\text{filter}}_{\text{token}}$ & 4.829 & 4.977 & 29.896 & 6.096 & 5.273 & 5.092 & 8.041 & 24.997 & 18.963 & 4.517 & 4.751 \\
\RatioShort\ (200 sample) & 3.094 & 3.109 & 3.929 & 3.084 & 2.885 & 3.017 & 3.029 & 4.095 & 3.603 & 2.891 & 2.991 \\
\RatioShort\ & 3.095 & 3.103 & 3.944 & 3.107 & 2.860 & 3.012 & 3.050 & 4.037 & 3.622 & 2.911 & 2.994 \\
\midrule
Post-Training Performance & 29.6 & 26.4 & 22.9 & 33.0 & 33.0 & 31.2 & 30.6 & 24.4 & 18.2 & 34.2 & 33.3 \\
\bottomrule
\end{tabular}%
}
\caption{Full metric assessment results on Qwen-2.5-3B.
Model names are abbreviated with Q for Qwen.}
\label{tab:full_qwen3}
\end{table*}

\newtcblisting{PromptBox}[1][]{%
  enhanced,
  breakable,
  colback=white,
  colframe=black!50,
  boxrule=0.6pt,
  arc=1mm,
  left=10pt,
  right=10pt,
  top=1.0mm,
  bottom=1.0mm,
  title=Evaluation prompt for LLM-judged quality assessment,
  fonttitle=\bfseries,
  listing only,
  listing options={
    basicstyle=\ttfamily\footnotesize, %
    breaklines=true,
    breakatwhitespace=false,
    columns=fullflexible,
    keepspaces=true,
    showstringspaces=false
  },
  #1
}

\clearpage

\begin{PromptBox}
You are a meticulous and highly critical evaluator of AI reasoning. Your primary goal is to identify and quantify subtle flaws, logical gaps, inefficiencies, and hidden assumptions. Do not default to a high score. Your starting assumption should be critical, and you must rigorously justify every point awarded.

First, please carefully read the following problem statement:
<Problem>
{question}
</Problem>

Now, please carefully read the following candidate's chain-of-thought reasoning:
<Reasoning>
{reasoning_to_evaluate}
</Reasoning>

When evaluating this reasoning, you must adhere to the following five key evaluation criteria and the scoring rubric below.

Scoring Guidelines and Calibration:
You must use the full 0.0 to 1.0 scale. Scores should not be clustered at the top. Use this rubric to anchor your scores:
1.0 (Exceptional/Flawless): Reserved for reasoning that is not only correct but also elegant, insightful, and comprehensive. It is perfectly structured and leaves no room for doubt. This score should be exceedingly rare.
0.8 - 0.9 (Excellent but Imperfect): The core reasoning is valid and well-supported, but there may be very minor, superficial issues (e.g., a trivial typo in a formula that doesn't affect the outcome, a slightly awkward phrasing). The conclusion is unaffected.
0.5 - 0.7 (Competent but Flawed): The reasoning is generally on the right track but contains noticeable and non-trivial flaws. Examples include: a minor factual error, a logical leap that requires the reader to fill in the blanks, an inefficient method where a much simpler one exists, or a partially incomplete answer.
0.2 - 0.4 (Poor): The reasoning contains fundamental flaws that largely invalidate the process or conclusion. Examples include: a significant factual error, a clear logical fallacy, misunderstanding of the core problem constraints.
0.0 - 0.1 (Unacceptable): The reasoning is completely incorrect, irrelevant, nonsensical, or makes no meaningful attempt to solve the problem.

Crucial Instruction for High Scores:
To combat score inflation, you must justify high scores with the same rigor as low scores. For any criterion where you assign a score of 0.9 or 1.0, your justification must explicitly state what makes the reasoning exceptional and why it lacks even subtle flaws.

Evaluation Criteria:
Factual Accuracy:
Scrutinize every claim, formula, and piece of domain knowledge. Is it precisely correct? Assess the application of problem constraints, paying close attention to edge cases and boundary conditions. Penalize any inaccuracy, no matter how small.

Logical Rigor:
Probe for hidden assumptions and unstated premises. Does each conclusion necessarily and unambiguously follow from the preceding steps? Identify any logical fallacies, contradictions, or jumps in reasoning. A chain is only as strong as its weakest link.

Solution Completeness:
Does the reasoning address all parts of the problem statement exhaustively? Does it consider all possible cases, sub-problems, and nuances? An answer that is correct for one case but ignores others is incomplete.

Reasoning Efficiency:
Is this the most direct and economical path to the solution? Penalize any unnecessary complexity, redundant steps, or exploration of irrelevant tangents, even if they eventually lead to the correct answer. The cognitive effort should be proportionate to the problem's complexity.

Presentation Quality:
How clearly is the reasoning communicated? Is the structure logical and easy to follow? Ambiguous language, poor organization, or a confusing sequence of steps should be penalized. An observer should be able to verify the reasoning process without difficulty.

For each of the five evaluation criteria, please give a score from 0.0 to 1.0 (in 0.1 increments) and a brief, clear justification for that score in the JSON structure.

Your output must be a single, valid JSON object. The format of the JSON object is as follows:
```json
{
  "dimensional_evaluation": {
    "factual_accuracy": {
      "score": <float between 0.0 and 1.0>,
      "reason": "<Your justification for the factual accuracy score>"
    },
    "logical_rigor": {
      "score": <float between 0.0 and 1.0>,
      "reason": "<Your justification for the logical rigor score>"
    },
    "solution_completeness": {
      "score": <float between 0.0 and 1.0>,
      "reason": "<Your justification for the solution completeness score>"
    },
    "reasoning_efficiency": {
      "score": <float between 0.0 and 1.0>,
      "reason": "<Your justification for the reasoning efficiency score>"
    },
    "presentation_quality": {
      "score": <float between 0.0 and 1.0>,
      "reason": "<Your justification for the presentation quality score>"
    }
  },
  "overall_score": <float between 0.0 and 1.0>,
  "overall_reason": "<A concise summary justifying the overall score by synthesizing the key findings from the dimensional evaluation.>"
}
\end{PromptBox}

{\captionsetup{type=table}
\captionof{table}{Evaluation prompt for LLM-judged quality assessment as a baseline metric.}
\label{box:prompt}}

\end{document}